\documentclass{article}
\usepackage{graphicx}
\usepackage{amsthm}
\usepackage{cite}
\usepackage{amsmath,amssymb,amsfonts,bbm}
\usepackage{float}
\usepackage{graphicx}
\usepackage{textcomp}
\usepackage{xcolor}
\usepackage{bm}
\usepackage{subfigure} 
\usepackage{algorithm, algpseudocode}
\usepackage{booktabs}

\usepackage{iclr2021_conference,times}
\usepackage{hyperref}
\usepackage{url}

\newcommand{\tb}[2]{\begin{tabular}{@{}#1@{}}#2\end{tabular}}  

\newcommand{\eps}{{\varepsilon}}
\newcommand{\squishlist}{
\begin{list}{$\bullet$}
	{ \setlength{\itemsep}{0pt}      \setlength{\parsep}{3pt}
		\setlength{\topsep}{3pt}       \setlength{\partopsep}{0pt}
		\setlength{\leftmargin}{1.5em} \setlength{\labelwidth}{1em}
		\setlength{\labelsep}{0.5em} } }
\newcommand{\squishend}{
\end{list}  }

\title{Enjoy Your Editing: Controllable GANs for Image Editing via Latent Space Navigation}

\author{Peiye Zhuang, Oluwasanmi Koyejo, Alexander G. Schwing\\
University of Illinois at Urbana-Champaign\\
\texttt{\{peiye, sanmi, aschwing\}@illinois.edu} \\
}

\algnewcommand{\Input}{\item[\textbf{Input:}]}%
\algnewcommand{\Output}{\item[\textbf{Output:}]}
\algnewcommand{\RETURN}{\item[\textbf{Return:}]}

\begin{document}
\maketitle
\begin{abstract}
Controllable semantic image editing enables a user to change entire image attributes with a few clicks, e.g., gradually making a summer scene look like it was taken in  winter.
Classic approaches for this task use a Generative Adversarial Net (GAN) to learn a latent space and suitable latent-space transformations. However, current approaches often suffer from attribute edits that are entangled, global image identity changes, and diminished photo-realism.
To address these concerns, we learn multiple attribute transformations simultaneously, integrate attribute regression into the training of transformation functions, and apply a content loss and an adversarial loss that encourages the maintenance of image identity and photo-realism. We propose quantitative evaluation strategies for measuring controllable editing performance, unlike prior work, which primarily focuses on qualitative evaluation. Our model permits better control for both single- and multiple-attribute editing while preserving image identity and realism during transformation. We provide empirical results for both natural and synthetic images, highlighting that our model achieves state-of-the-art performance for targeted image manipulation. 
\end{abstract}

\vspace{-0.4cm}
\section{Introduction}
\vspace{-0.2cm}

Semantic image editing is the task of transforming a source image to a target image while modifying desired semantic attributes, e.g., to make an image taken during summer look like it was captured in winter. 
The ability to semantically edit images is useful for various real-world tasks, including artistic visualization, design, photo enhancement, and targeted data augmentation. 
To this end, semantic image editing has two primary goals:
(i)  providing continuous manipulation of multiple attributes simultaneously and (ii) maintaining the original image's identity as much as possible while ensuring photo-realism. 

Existing GAN-based approaches for semantic image editing can be  categorized roughly into two groups: 
(i) {\em image-space editing} methods  directly transform one image to another across domains~\citep{isola2017image, zhu2017toward, zhu2017unpaired, wu2019relgan, choi2018stargan, choi2019stargan, lee2020drit++}, usually using variants of generative adversarial nets (GANs)~\citep{goodfellow2014generative}. 
These approaches often have high computational cost, and they primarily focus on binary attribute (on/off) changes, rather than providing continuous attribute editing abilities. 
(ii) {\em latent-space editing} methods  focus on discovering latent variable manipulations that permit continuous semantic image edits. The chosen latent space is most often  the  latent space of GANs.
Both unsupervised and (self-)supervised latent space editing methods have been proposed. Unsupervised latent-space editing methods~\citep{voynov2020unsupervised, harkonen2020ganspace} are often less effective at providing semantically meaningful directions and all too often change image identity during an edit. Current (self-)supervised methods~\citep{jahanian2019steerability, plumerault2020controlling} are limited to geometric edits such as rotation and scale.
To our knowledge, only one supervised approach has been proposed~\citep{shen2019interpreting} -- developed to discover semantic latent-space directions for binary attributes. As we show, this method suffers from entangled attributes and often does not preserve image identity during manipulation.

\begin{figure}
		\vspace{-0.5cm}
	\centering
	\includegraphics[width=\linewidth]{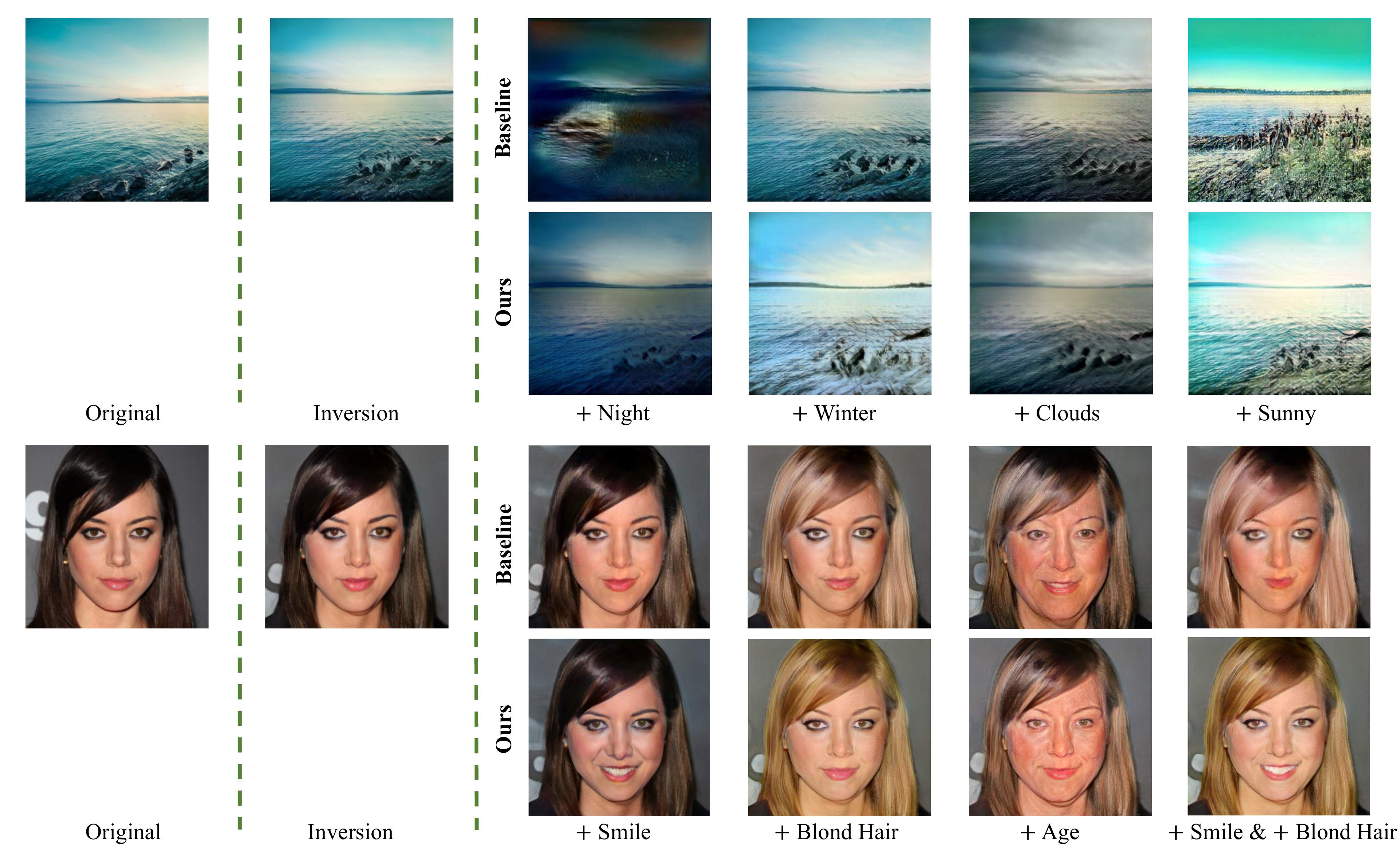}
		\vspace{-0.6cm}
	\caption{\textbf{Real image manipulation} on scene (top two rows, photo from Flickr) and face (bottom two rows, unseen image from CelebA-HQ) using pretrained StyleGAN2~\citep{karras2019analyzing}: 
	We reconstruct the real images (col.1)  by finding a latent vector with the best inversion result (col.2) on StyleGAN2~\citep{abdal2019image2stylegan}. After that, we transform the latent vectors for single- and multiple-attribute manipulations (col.3-6). Note that unlike ours, the baseline method~\citep{shen2019interpreting} either changes image identity or confounds semantic properties, or both.} 
	\vspace{-0.4cm}
	\label{fig:real_edit}
\end{figure}

{\bf Contributions.}
We propose a latent-space editing framework for semantic image manipulation 
that fulfills the aforementioned primary goals.
Specifically, we use a GAN and employ a joint sampling strategy trained to edit multiple attributes simultaneously. 
To disentangle attribute transformations in the latent space of GANs, we integrate a regressor to predict the attributes that an image exhibits. 
The regressor also permits precise control of the manipulation degree and is easily extended to multiple attributes simultaneously.
In addition, we incorporate a perceptual loss~\citep{li2019controllable} and an adversarial loss that helps preserve image identity and photo-realism during manipulation. 

We compare our method to several popular frameworks, from existing image-to-image translation methods~\citep{zhu2017unpaired, wu2019relgan, choi2019stargan} to latent space transformation-based approaches~\citep{shen2019interpreting, voynov2020unsupervised}.
We mention that prior work primarily uses qualitative evaluation like the one in Fig.~\ref{fig:real_edit}. In contrast, we propose a quantitative evaluation  to measure {\em controllability}. 
Both qualitative and quantitative results provide evidence that our approach outperforms prior work in terms of quality of the semantic image manipulation while maintaining image identity. 

\vspace{-0.1cm}
\section{Related work}
\label{sec:rel}
\textbf{Generative Adversarial Networks (GANs)}~\citep{goodfellow2014generative} have significantly improved realistic image generation in recent years~\citep{ karras2017progressive, jolicoeur2018relativistic, zhang2018self, brock2018large, Park_2019_CVPR, karras2019style, karras2019analyzing}. For this, a GAN formulates a 2-player non-cooperative game between two deep nets: (i) a generator that produces an image given a random noise vector in the latent space, sampled from a known prior distribution, usually a normal or a uniform distribution; (ii) a discriminator whose input is both synthetic and real data, which is to be differentiated. 

\textbf{Semantic image editing} seeks to automate image manipulation of semantics. Encouraged by the success of deep nets, recent works have applied deep learning methods for semantic image editing tasks such as style transfer~\citep{luan2017deep, li2017universal}, image-to-image translation~\citep{zhu2017unpaired, isola2017image,   wang2018high, wu2019relgan, choi2018stargan, choi2019stargan, lee2020drit++} and discovering semantically meaningful directions in a GAN latent space~\citep{jahanian2019steerability, plumerault2020controlling,shen2019interpreting, voynov2020unsupervised, harkonen2020ganspace}. Note that our task is an extended version of semantic image editing that requires more comprehensive control to satisfy user-desired operations. Therefore, most aforementioned approaches do not meet the requirements.
Nonetheless, we categorize the approaches most relevant  to our task into two groups: (i) image editing via manipulation in image space, and (ii) image editing via latent space navigation. 

\textbf{Image-space editing using GANs} directly manipulates an image for targeted editing. 
Early work~\citep{isola2017image} used GANs to implement semantic translations between two image domains with paired data, e.g., day to night. Follow-up work focused on multi-modal~\citep{wang2018high, wu2019relgan, choi2018stargan, choi2019stargan, lee2020drit++, bhattarai2020inducing} and unpaired image domain translation~\citep{zhu2017unpaired}. In this case, they primarily consider binary (on/off) attribute changes, regardless of whether the process is dynamic, for example, day to night.

\textbf{Latent-space editing via GANs} has received an increasing amount of recent interest. Most prior work focused on identifying semantically meaningful directions in the latent space of GANs, so that shifting latent vectors towards these directions achieves the desired image manipulation. Recent papers found semantics in the latent space of GANs, such as color transformations and camera movements~\citep{jahanian2019steerability}, or face attribute changes~\citep{shen2019interpreting}, such as smile. Other work considered unsupervised methods~\citep{harkonen2020ganspace, voynov2020unsupervised} to discover interpretable latent space directions. 
However, additional challenges inherent to unsupervised manipulation of the latent space 
arise, and have not been addressed in prior work, e.g., direction quality with regard to degree control and image identity preservation.
\vspace{-0.3cm}
\section{Method}
\vspace{-0.2cm}
\label{sec:method}
\subsection{Problem Statement}
\vspace{-0.1cm}
We consider controllable semantic image editing via latent space navigation in GANs.
We begin with a \textit{fixed} GAN model that consists of a generator $G$ and a discriminator $D$. The input of  $G$ is a latent vector $\bm z \in \mathbb{R}^m$ from a latent space $\mathcal{Z}$. Here, $m$ is the dimensionality of the latent space. Given $N$ attributes, we aim to discover semantically meaningful latent-space GAN directions, $\bm T = \{\bm d_1, \dots, \bm d_N\}$, to manipulate the attributes of synthetic images $G(\bm z)$ with an assigned step size $\bm \eps = \{\varepsilon_1, \dots, \varepsilon_N\}$, where $\bm d_i \in \mathbb{R}^m \text{ for all } i \in \{1,\dots, N\}$.
In the end, each attribute of an edited image can be changed with the corresponding degree $\bm \eps$ from $\bm \alpha = \{\alpha_1, \dots, \alpha_N\}$, the original attributes of  $G(\bm z)$ predicted by a regressor $R$.

\begin{figure}
	\centering
	\vspace{-0.3cm}
	\includegraphics[width=\linewidth]{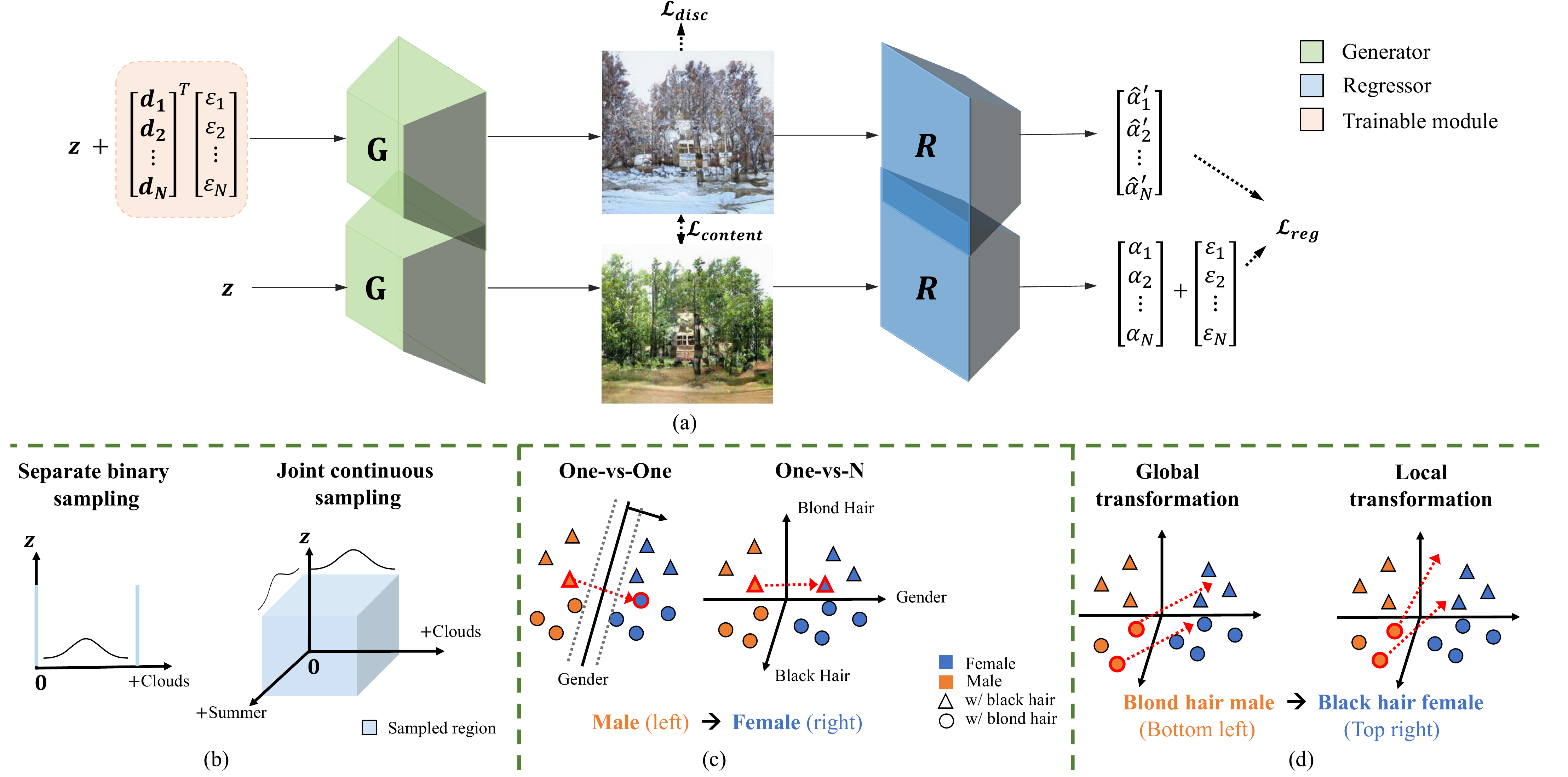}
	\vspace{-0.5cm}
	\caption{\textbf{Our overall framework} (top row) and  3 training strategies different from  prior work (bottom row). In (a), $G$ takes $\bm z$ and an edited latent vector separately to synthesize images.
  $\bm T = \{\bm d_1, ..., \bm d_N\}$ are the trainable latent-space directions and ${\bm \eps}$ represents transformation extent. 
  Original and edited image attributes, ${\bm \alpha}$ and $\bm {\hat{\alpha}'}$, are predicted by the pre-trained regressor $R$.
  The discriminator loss $\mathcal L_\text{disc}$ (the discriminator $D$ not shown due to limited space), the regression loss $\mathcal L_\text{reg}$, and the perceptual loss $\mathcal L_\text{content}$ are  used to update $\bm T$. We explain (b-d) in Sec.~\ref{approach}.
	  }
  \vspace{-0.4cm}
	\label{fig:model}
\end{figure}

\vspace{-0.1cm}
\subsection{Proposed approach}
\label{approach}
We provide an overview of our method in Fig.~\ref{fig:model} (a), and further illustrate how our approach differs from prior work
in Fig.~\ref{fig:model} (b-d).

\textbf{Overview.} 
As shown in Fig.~\ref{fig:model} (a), we employ a GAN model consisting of a generator $G$ and a discriminator $D$ (not shown due to limited space), as well as a regressor $R$, all of which are pre-trained. Our goal is to find latent directions $\bm T$ that provide attribute specific image edits. 
At each training step, $G$ takes $\bm z$ and an edited latent vector referred to as $\bm z'$. We follow prior work which suggests that direction vectors in latent space of GANs permit image editing~\citep{jahanian2019steerability}. Formally, given a transformation degree $\bm \eps$, we obtain the latent-space edit as $\bm z' = \bm z + \bm T \bm \eps = \bm z + \sum_{i=1}^{N}\eps_i \bm d_i$.
$G$ provides the recovered and edited images $G(\bm z)$ and $G(\bm z')$, which are separately processed by the regressor $R$  to predict attribute values. The original attributes of $G(\bm z)$ are $\bm \alpha  = R(G(\bm z))$. 
The (pseudo-) ground truth and predicted attribute of $G(\bm z')$ are $\bm \alpha'$ and  $\hat{\bm \alpha}'$, respectively, where $\bm \alpha' = \bm \alpha + \bm \eps$ and $\hat{\bm  \alpha}' = R(G(\bm z'))$. 

Intuitively, the goal of $\bm T$ is to transform $\bm z$ by adding semantically meaningful information such that the corresponding output image $G({\bm z'})$ exhibits attribute changes  $\bm \eps$ from $\bm \alpha$. 
 In practice, we normalize the range of attribute values to $[\bm 0, \bm 1]$, i.e., both $\bm \alpha$ and $\bm \alpha'\in [\bm 0, \bm 1]$. During training, we maintain the unit range by controlling the given $\bm \eps$.
Formally, $\bm \eps$ is drawn from a distribution $\mathcal{D}_\eps$ uniform in $[-1, 1]^N$ while considering the constraint that $\bm 0 \leq \bm \alpha + \bm \eps \leq \bm 1$.

\textbf{Objective function.} To find $\bm T$ we minimize the weighted objective:
\begin{align}
\min_{\bm T} \mathcal{L} \triangleq \lambda_1 \mathcal{L}_\text{reg} + \lambda_2 \mathcal{L}_\text{disc} + \lambda_3\mathcal{L}_\text{content}.
\vspace{0.1cm}
\label{eq:4}
\end{align}
Note, the objective is only used for optimizing $\bm T$, while the other modules remain fixed. Hyperparameters $\lambda_1, \lambda_2,$ and $\lambda_3$ adjust the loss terms. 

The regression loss $\mathcal{L}_\text{reg}$ assesses whether $\bm T$ performs the transformations indicated by $\bm \eps$. We express the regression loss via a binary cross entropy: 
\begin{equation}
\mathcal{L}_\text{reg} =\quad \mathbb{E}_{\bm z \sim \mathcal{Z}, \bm \eps \sim \mathcal{D}_\eps} 
[-\bm{\hat{\alpha}'} \log{\bm \alpha'} - (\bm 1-\bm{\hat{\alpha}'}) \log{(\bm 1-\bm \alpha')}].
\end{equation}
Note that $\bm \alpha'$ is from the distribution generated by $\bm z \text{ and } \bm \eps$. Please refer to the appendix C.1 for more details on this distribution.

The second loss term $\mathcal{L}_\text{disc}$ is computed using the discriminator $D$ and measures the quality of the  generated images, i.e., 
\begin{align}
\mathcal{L}_\text{disc} = \quad \mathbb{E}_{\bm{z'} \sim \mathcal{Z'}|\bm z} \left[\log (  1-D(G(\bm{z'}))) \right].
\end{align}
Here we refer to the domain of $\bm z'$ via $\mathcal{Z}'$, which is conditioned on $\bm z$. We write this using $\mathcal{Z}'|\bm z$.

Lastly, we use a content loss $\mathcal{L}_\text{content}$, often also referred to as perceptual loss. It is designed to estimate the distance between two images and it is employed to maintain the image identity during the transformation. 
Specifically, we use the content loss term   
\begin{align}
&\mathcal{L}_\text{content} =
\quad \mathbb{E}_{\bm{z} \sim \mathcal{Z}, \bm{z'} \sim \mathcal{Z'}|\bm z} \sum_{i \in \mathcal{D}_\text{content}}  \|F_i(G(\bm{z}')) - F_i(G(\bm{z}))\|^2_2,
\end{align}
where $F_i(\cdot)$ denotes a feature function which extracts intermediate features from images. $\mathcal D_\text{content}$ indicates the layers of a pre-trained model which are used as features. We approximate the aforementioned expectations by empirical sampling. We defer  algorithm details to Appendix A.

\textbf{Joint-distribution sampling and training.} 
The regressor operates in a multi-label setting, i.e., each data possesses multiple attributes. To ensure the disentanglement of attribute edits we sample synthetic images from the entire data distribution and find all transformations at once (illustrated in Fig.~\ref{fig:model} (b,c) right). In contrast, ~\citet{shen2019interpreting} prepare training samples on the two opposing data subsets with regards to an attribute, e.g., no clouds \emph{vs}.\ many clouds, and find the directions with $N$ one-vs-one classifiers (sketched in Fig.~\ref{fig:model} (b,c) left).

\textbf{Transformation module $\bm T$.} We study two types of transformations $\bm T$:
(i) \textit{global} and (ii) \textit{local}. 
A \textit{global} transformation $\bm T$ refers to a semantic latent-space transformation identical for all $\bm z$ during inference. This is illustrated via parallel red dashed arrows in Fig.~\ref{fig:model} (d) left. These global directions are commonly used~\citep{jahanian2019steerability, shen2019interpreting, viazovetskyi2020stylegan2, harkonen2020ganspace}. However, a globally identical direction might not serve all data.
In contrast, a \textit{local} transformation $\bm T$ is a function of $\bm z$ which provides various directions for different $\bm z$ (in Fig.~\ref{fig:model} (d) right, shown via non-parallel red arrows). Formally, $\bm d_i = f_\theta^i(\bm z)$, where $f_\theta^i$ is implemented via a deep net. In Sec.~\ref{sec:exp} we show that the local transformation $\bm T$ succeeds on failure cases of the global transformation.

\begin{figure}[t]
	\centering
	\includegraphics[width=\linewidth]{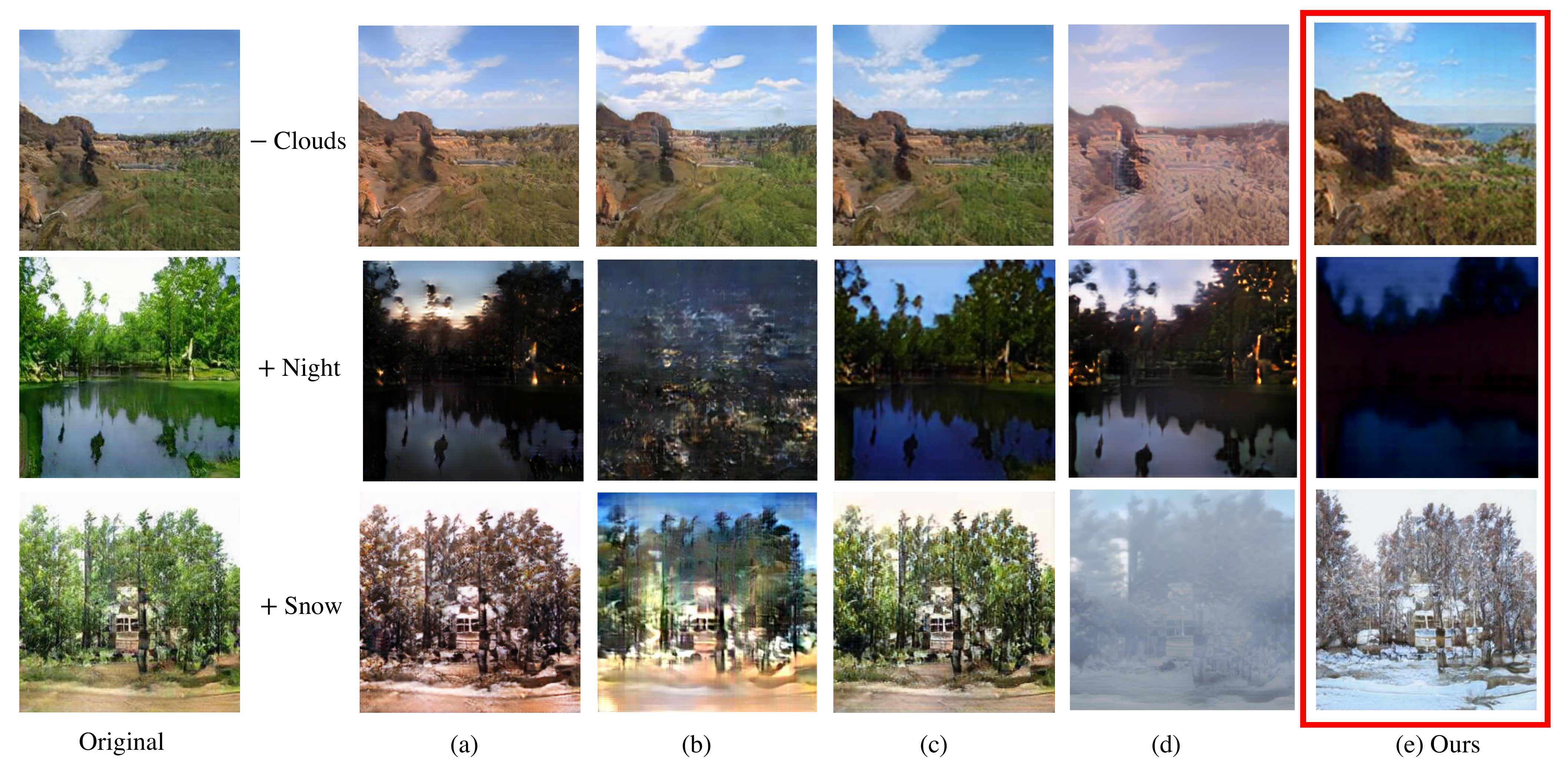}
	\caption{\textbf{Comparison of  image-space editing approaches} with respect to ``removing clouds'' (top), ``enhancing night'' (middle) and ``adding snow'' (bottom). The original images (col.1) are followed by a given editing task. From (a-e): (a) CycleGAN~\citep{zhu2017unpaired}, (b) StarGAN v2~\citep{choi2019stargan}, (c) RelGAN~\citep{wu2019relgan}, (d) DRIT++~\citep{lee2020drit++}, and (e) Ours.}	\label{fig:comparison}
\end{figure}

\begin{figure}
	\centering
	\includegraphics[width=\linewidth]{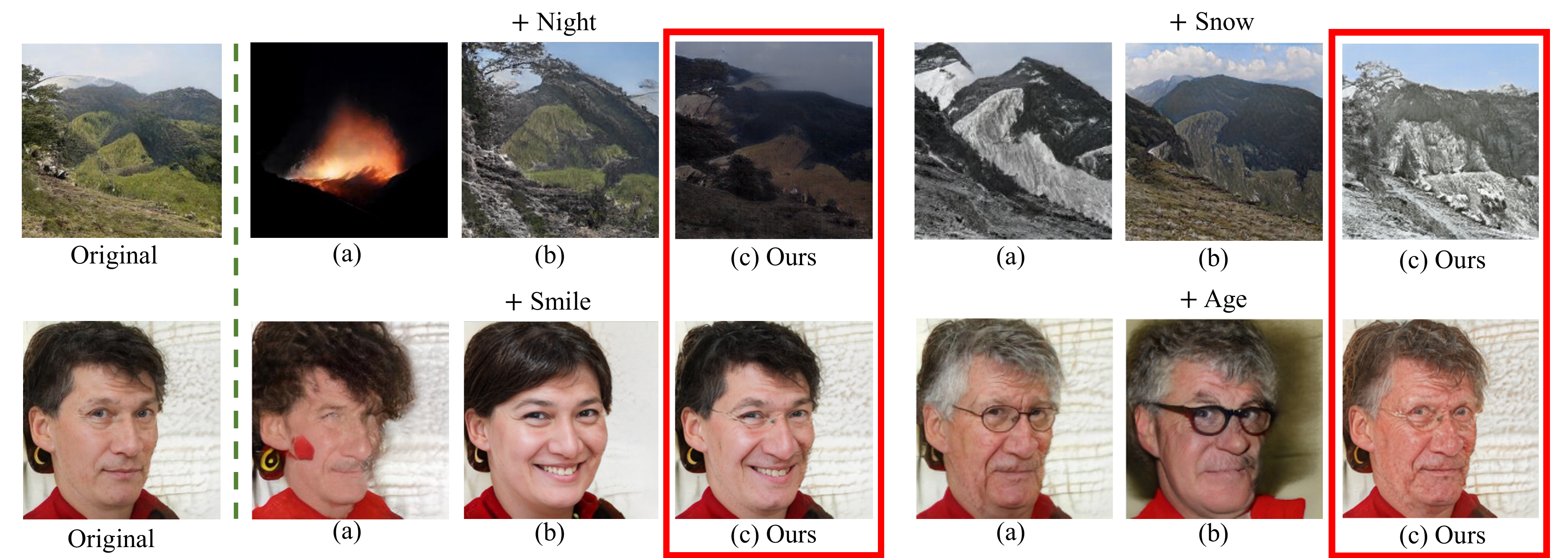}
	\caption{\textbf{Comparison of latent-space editing approaches on StyleGAN2.} (a) ~\citet{shen2019interpreting}; (b)~\citet{voynov2020unsupervised}; (c) Ours. Original synthetic image (col.1); edited results of a semantic manipulation noted on top (col.2-4; col.5-7).  \citet{shen2020closedform} mention that ``age,'' ``gender,'' and ``glasses'' directions are hard to disentangle potentially due to data bias, while our results suggest a better direction disentanglement.}
	\label{fig:latent_comp}
\end{figure}

\section{Experiments}
\label{sec:exp}
\vspace{-0.2cm}
We introduce datasets, implementation details, evaluation metrics, and show results  next.

\textbf{Datasets.} We evaluate the proposed approach on two types of datasets: (i) face datasets -- FFHQ~\citep{karras2019style}, CelebA~\citep{liu2018large} and CelebA-HQ~\citep{karras2017progressive}, commonly used in prior work~\citep{karras2017progressive, karras2019style, karras2019analyzing, shen2019interpreting, harkonen2020ganspace, viazovetskyi2020stylegan2}. (ii) natural scene datasets -- Transient Attribute Database~\citep{laffont2014transient} and MIT Places2 data~\citep{zhou2017places}, which contain attributes suitable for continuous semantic image editing. We briefly introduce the scene datasets:
\squishlist
\item {\em Transient Attribute Database}~\citep{laffont2014transient}: It contains 8,571 scene images with 40 attributes in 5 categories including lighting (e.g., ``bright''), weather (e.g., ``cloudy'') , seasons (e.g., ``winter''), subjective impressions (e.g., ``beautiful''), and additional attributes (e.g., ``dirty'').
Each attribute is annotated with a real-valued score between $0$ and $1$, where $0$ indicates absence of the attribute.

\item {\em MIT Places2 data}~\citep{zhou2017places}: Using the provided category annotations (i.e., indoor/outdoor and natural/artificial), we select the natural outdoor scenes, obtaining a total of 144,543 images. 

\squishend

\label{sec:eval_detail}
\textbf{Implementation details.} We choose $\lambda_1=10, \lambda_2 = 0.05, \lambda_3= 0.05$ in Eq.~\eqref{eq:4} and compute the perceptual loss using the   $\text{conv}1\_2, \text{conv}2\_2, \text{conv}3\_2, \text{conv}4\_2$ activations in a VGG-19 network~\citep{simonyan2014very} pre-trained on the ImageNet dataset~\citep{russakovsky2015imagenet}. We train $\bm T$ for 50k iterations with a batch size of 4. An Adam optimizer is used with a learning rate of $1\textrm{e-4}$. For the latent space ${\cal Z}$ in the proposed method we use the $\mathcal{W}$ space of StyleGAN2 and the $\mathcal{Z}$ space of PGGAN. 
For the attribute regressor $R$, we adopt a ResNet-50~\citep{he2016deep} for attribute prediction on the Transient Attribute~\citep{laffont2014transient} and the CelebA~\citep{liu2018large} data.
The last fully connected layer in the ResNet-50 is replaced by a linear layer with an output dimension of 40. We train the regressors for 500 epochs and use the weights with the best validation mean squared error (MSE) on CelebA and the best test MSE on the Transient Scene Database. The GAN follows the StyleGAN2 architecture and is pretrained with 200k iterations on a union of the two natural scene datasets using a resolution of $256 \times 256$.
The training batch size is 32 and an Adam optimizer is used with a learning rate of $2\textrm{e-3}$. For the face datasets, we use FFHQ pre-trained weights\footnote{https://github.com/rosinality/stylegan2-pytorch} of StyleGAN2, and CelebA-HQ weights of PGGAN\footnote{https://pytorch.org/hub}.

\textbf{Baselines.} We compare our approach to several popular image-to-image translation approaches~\citep{zhu2017unpaired, wu2019relgan, choi2019stargan} and latent space direction discovery methods ~\citep{shen2019interpreting,voynov2020unsupervised}. Since~\citet{zhu2017unpaired, choi2019stargan, shen2019interpreting} cannot deal with continuous attributes, we split the data into 2 domains using a threshold value of $0.5$ for each of the 40 attributes. Afterward, the models are trained on 2 contrast domains, e.g., with or without a certain attribute. We use the official code of all baseline methods and rigorously follow their training steps. Note that work by~\citet{voynov2020unsupervised} is  unsupervised, i.e., the latent-space directions are human interpreted. To avoid bias during the selection of directions, we use the attribute regressor to automatically identify the most significant directions that can edit predetermined attributes. The details of preparing the baselines are given in Appendix~C.2.

\textbf{Evaluation metrics.} There is no good numerical metric to evaluate image editing~\citep{shen2019interpreting, voynov2020unsupervised}. In an attempt to address this concern, we automate quantitative evaluation based on a property that editing of attributes should maintain image identity. To achieve this, we employ a popular image identity recognition model\footnote{https://github.com/ox-vgg/vgg\_face2} pre-trained on the VGGface2 data~\citep{Cao18}. Cosine similarity is used to represent
the similarity between paired original and edited images~\citep{Cao18}. In addition, we evaluate changes of the other independent attributes using the pre-trained regressor.

\begin{figure}[t]
	\vspace{-0.2cm}
	\includegraphics[width=\linewidth]{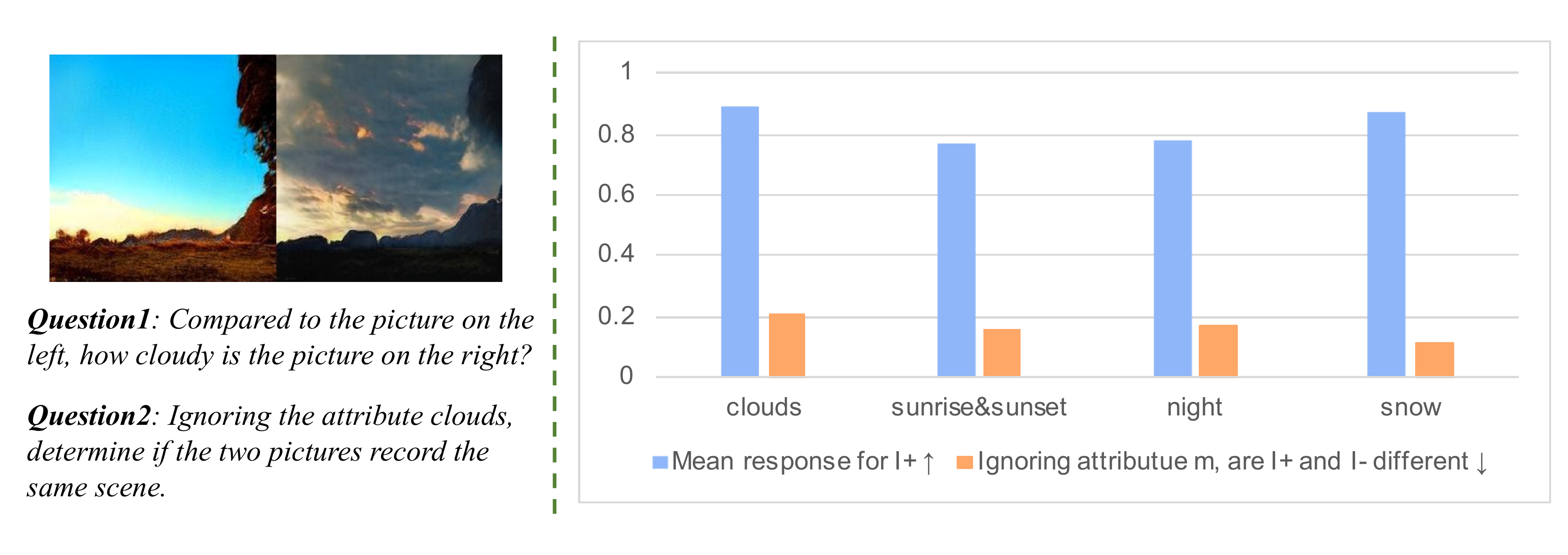}
	\vspace{-0.6cm}
	\caption{\textbf{User study for image editing and image identity preservation}. Blue bars show the predicted presence of a given attribute for images ($I+$, higher is better).  The orange bars measure the difference between paired images (lower is better). }
	\label{fig:userstudy}
\end{figure}

\subsection{Results on natural scene datasets} 

\textbf{Comparison to image-to-image translation:} As shown in  Fig.~\ref{fig:comparison} (a-d), 
we observe image-to-image translation to perform poorly when editing image details (e.g., removing clouds). Moreover, sometimes  artifacts (orange spots in (a) and (b)) are introduced.
In contrast, our model shows overall compelling transfer performance on all these attributes. 

\textbf{Comparison to latent-space translation:} Inspecting Fig.~\ref{fig:real_edit}, we observe  our approach to improve upon work by \citet{shen2019interpreting} in attribute edit quality and number of artifacts.
Similarly, in Fig.~\ref{fig:latent_comp}, we notice that our method performs well in attribute editing, and has a better ability to preserve image identity when adding ``night'' and ``snow.'' We show additional analysis and results in  Appendix D and E.1.

\begin{figure}[t]
	\includegraphics[width=\linewidth]{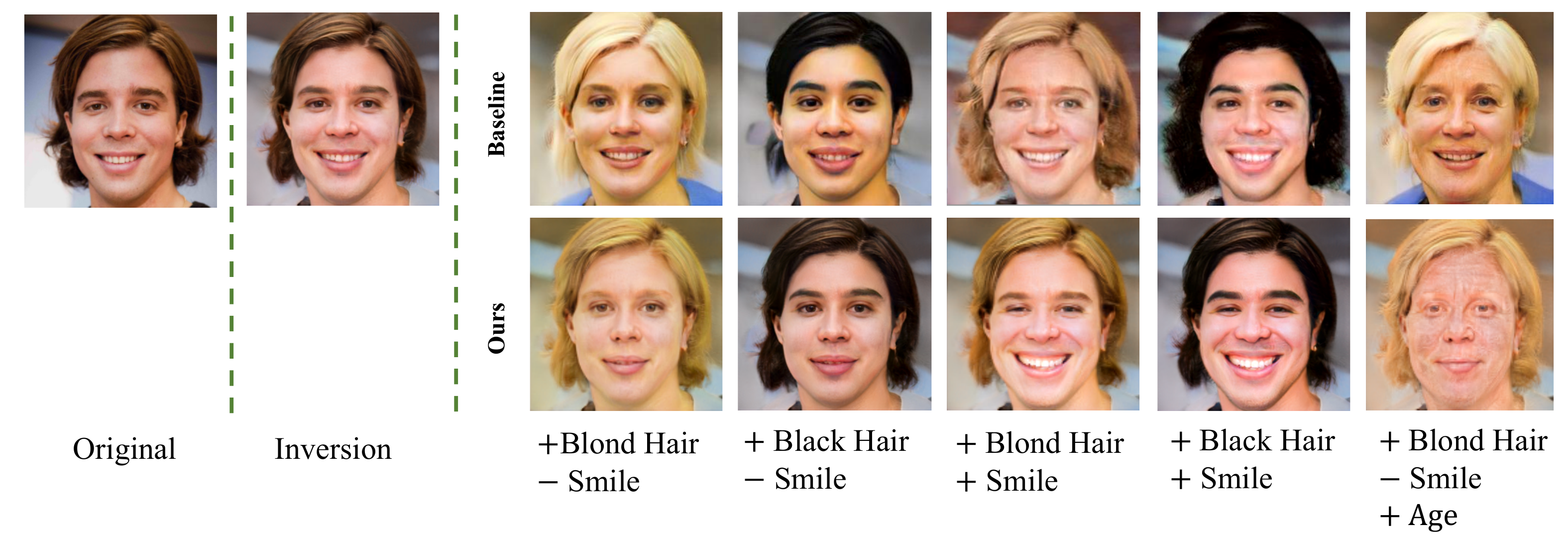}
	\vspace{-0.6cm}
	\caption{\textbf{Real image synthesis with multiple attributes.} ~\citet{shen2019interpreting} (top row); ours (bottom row): original real image (col.1); inverted result (col.2) from \citet{abdal2019image2stylegan}; edited results (col.3-7) with target attribute changes shown at the bottom. }
	\label{fig:comb_face_attrs}
	\vspace{-0.3cm}
\end{figure}

\textbf{Quantitative evaluation:} We follow a user study proposed by~\citet{kowalski2020config} to evaluate  controllability of our scene edits.
Concretely, given an image $I = G(\bm z)$, we generate paired images ($I+$, $I-$) that have opposing values for attribute $n$. 40 users took the test to answer: (a) the presence and strength of the attribute $n$ in $I+$, and (b) if $I+$ and $I-$ are  identical when ignoring the attribute. Both types of questions have options on a 5-level scale from $0$ (``not at all'') to $1$ (``totally''). Ideally, the response to question (a) should be $1$ while we expect replies for  (b) to be $0$. We test performance of image editing with ``clouds'', ``sunrise\&sunset'', ``night'' and ``snow'' attributes.
A question example and statistical results over 50 image pairs are shown in Fig.~\ref{fig:userstudy}. Statistical evidence supports our claim that the proposed approach performs well with regards to image edits while maintaining image identity.

\begin{table}[h]
\centering
\setlength{\tabcolsep}{3.5 pt}{
\begin{tabular}{cccccccccc}
\toprule
& \multicolumn{3}{c}{Smile}               
& \multicolumn{3}{c}{Hair color}          
& \multicolumn{3}{c}{Smile $+$ Hair color}  \\ \midrule
$\vert \hat{\eps} \vert$ 
 & {(0, .3]}   & {(.3, .6]}  & {(.6, .9]} 
 & {(0, .3]}   & {(.3, .6]}  & {(.6, .9]} 
 & {(0, .3]}   & {(.3, .6]}  & {(.6, .9]}\\ \midrule
 
{Shen et al.} & \tb{c}{.202\\ {\small$\pm$ 5$e$-2}} 
        & \tb{c}{.204 \\ {\small$\pm$ 6$e$-2}}  
        & \tb{c}{ .224 \\{\small$\pm$ 4$e$-2}}
        & \tb{c}{ .256 \\{\small$\pm$ 1$e$-2}} 
        & \tb{c}{ .272 \\ {\small$\pm$ 2$e$-2}} 
        & \tb{c}{ .277 \\ {\small$\pm$ 2$e$-2}}
        & \tb{c}{ .299 \\{\small$\pm$ 5$e$-3}}
        & \tb{c}{.318 \\ {\small$\pm$ 1$e$-2}}
        & \tb{c}{.329 \\ {\small$\pm$ 3$e$-2}} \vspace{3pt} \\ 
{Voynov et al.} & \tb{c}{.115 \\ {\small$\pm$ 6$e$-3}}
        & \tb{c}{.211 \\ {\small$\pm$ 4$e$-2}} 
        & \tb{c}{.277 \\ {\small$\pm$ 7$e$-3}}
        & \tb{c}{.162 \\ {\small$\pm$ 1$e$-2}} 
        & \tb{c}{.166 \\ {\small$\pm$ 3$e$-2}} 
        & \tb{c}{.177 \\ {\small$\pm$ 9$e$-3}}
        & \tb{c}{.155 \\ {\small$\pm$ 4$e$-3}}  
        & \tb{c}{.220 \\ {\small$\pm$ 4$e$-2}}  
        & \tb{c}{.284 \\ {\small$\pm$ 2$e$-2}} \vspace{3pt} \\ 
Ours    & \tb{c}{$\bm{.085}$ \\{\small$\pm$ 4$e$-2}} 
        & \tb{c}{$\bm{.084}$ \\ {\small$\pm$ 3$e$-3}}
        & \tb{c}{$\bm{.098}$ \\ {\small$\pm$ 4$e$-3}}  
        & \tb{c}{$\bm{.075}$ \\ {\small$\pm$ 7$e$-4}} 
        & \tb{c}{$\bm{.083}$ \\ {\small$\pm$ 5$e$-3}} 
        & \tb{c}{$\bm{.084}$ \\ {\small$\pm$ 5$e$-3}} 
        & \tb{c}{$\bm{.088}$ \\ {\small$\pm$ 7$e$-3}} 
        & \tb{c}{$\bm{.111}$ \\ {\small$\pm$ 4$e$-3}} 
        & \tb{c}{$\bm{.134}$ \\ {\small$\pm$ 4$e$-3}}\\ 
\bottomrule
\end{tabular}}
\caption{\textbf{Quantitative evaluation of numerical changes on the other semantically independent attributes} (lower is better). (\textit{first row}) edited attributes; (\textit{second row}) the absolute changing range of the edited attributes; (\textit{bottom three rows}) averages (up) and standard deviations (down) in each row.}
\label{tabel:attr}
\end{table}

\begin{table}[h]
\vspace{-0.1cm}
\centering
\setlength{\tabcolsep}{3 pt}{
\begin{tabular}{cccccccccc}
\toprule
& \multicolumn{3}{c}{Smile}               
& \multicolumn{3}{c}{Hair color}          
& \multicolumn{3}{c}{Smile $+$ Hair color}  \\ \midrule
$\vert \hat{\eps} \vert$ 
 & {(0, .3]}   & {(.3, .6]}  & {(.6, .9]} 
 & {(0, .3]}   & {(.3, .6]}  & {(.6, .9]} 
 & {(0, .3]}   & {(.3, .6]}  & {(.6, .9]} \\ \midrule
{Shen et al.}  
        & \tb{c}{.918 \\ {\small$\pm$ 4$e$-3}} 
        & \tb{c}{.916 \\ {\small$\pm$ 1$e$-3}} 
        & \tb{c}{.907 \\ {\small$\pm$ 5$e$-3}}
        & \tb{c}{.887 \\ {\small$\pm$ 3$e$-2}} 
        & \tb{c}{.877 \\ {\small$\pm$ 4$e$-2}} 
        & \tb{c}{.874 \\ {\small$\pm$ 4$e$-2}}
        & \tb{c}{.877 \\ {\small$\pm$ 6$e$-3}} 
        & \tb{c}{.819 \\ {\small$\pm$ 3$e$-2}} 
        & \tb{c}{.801 \\ {\small$\pm$ 4$e$-2}} \vspace{3pt}  \\
{Voynov et al.} 
        & \tb{c}{.979 \\ {\small$\pm$ 2$e$-3}} 
        & \tb{c}{.896 \\ {\small$\pm$ 4$e$-3}} 
        & \tb{c}{.869 \\ {\small$\pm$ 9$e$-3}} 
        & \tb{c}{.955 \\ {\small$\pm$ 6$e$-2}} 
        & \tb{c}{.909 \\ {\small$\pm$ 4$e$-2}} 
        & \tb{c}{.904 \\ {\small$\pm$ 3$e$-2}} 
        & \tb{c}{.940 \\ {\small$\pm$ 9$e$-3}} 
        & \tb{c}{.829 \\ {\small$\pm$ 4$e$-2}} 
        & \tb{c}{.811 \\ {\small$\pm$ 3$e$-2}} \vspace{3pt} \\
Ours    & \tb{c}{$\bm{.995}$\\ {\small$\pm$ 7$e$-4}}  
        & \tb{c}{$\bm{.994}$\\ {\small$\pm$ 1$e$-3}} 
        & \tb{c}{$\bm{.992}$\\ {\small$\pm$ 9$e$-4}} 
        & \tb{c}{$\bm{.993}$\\ {\small$\pm$ 5$e$-4}} 
        & \tb{c}{$\bm{.993}$\\ {\small$\pm$ 1$e$-4}} 
        & \tb{c}{$\bm{.993}$\\ {\small$\pm$ 2$e$-4}} 
        & \tb{c}{$\bm{.992}$\\ {\small$\pm$ 9$e$-4}} 
        & \tb{c}{$\bm{.986}$\\ {\small$\pm$ 2$e$-3}} 
        & \tb{c}{$\bm{.984}$\\ {\small$\pm$ 4$e$-4}} \\ \bottomrule
\end{tabular}}
\caption{\textbf{Quantitative evaluation on image identity preservation} (higher is better). Notation is identical to Tab.~\ref{tabel:attr}.}
\label{tabel:id}
\end{table}
\subsection{Results on face datasets} 
\textbf{Manipulation results on StyleGAN2:}
The comparison on real images shown in Fig.~\ref{fig:real_edit} suggests that our method works well for  attribute edits.
Further, the synthetic image edit results in Fig.~\ref{fig:latent_comp} indicate that our edits are disentangled, while the baselines unexpectedly add ``glasses'' when aging the face. To increase the task difficulty, we edit real images with multiple attribute changes simultaneously. Results are summarized in Fig.~\ref{fig:comb_face_attrs}, which highlights the controllability of our edits.

\textbf{Quantitative evaluation:} We measure the changing degrees of the other independent attributes and the image identity when editing  attributes with various degrees $\hat \eps$. Here $\hat \eps$ is a predicted changing degree on a target attribute by $R$.
For a thorough comparison, we evaluate the performance on 3 segments according to the absolute value of $\hat \eps$, i.e., $\vert \hat{\eps}\vert$ in the range of $(0, 0.3], (0.3, 0.6] \text{ and } (0.6, 0.9]$. 
Evaluation on ``smile,'' ``hair color,'' and ``smile+hair color'' attributes are shown in Tab.~\ref{tabel:attr} and Tab.~\ref{tabel:id}. The ``hair color'' attribute includes ``blond'' and ``black'' colors where we average the results on both cases. For  multiple attribute editing, ``smile + hair color,'' we evaluate the case that both the two targeted attributes change within the $\vert \hat{\eps}\vert$ range. We use around 1k original images, generate 10k edited images with regard to each target attribute, and repeat the experiment 3 times. The averaged results and the standard deviations presented in Tab.~\ref{tabel:attr} and Tab.~\ref{tabel:id} suggest that our model outperforms the baselines with regard to disentanglement and image identity preservation. 

\begin{figure}
 \centering
\subfigure{
\begin{minipage}{7cm}
\centering
\includegraphics[width=\linewidth]{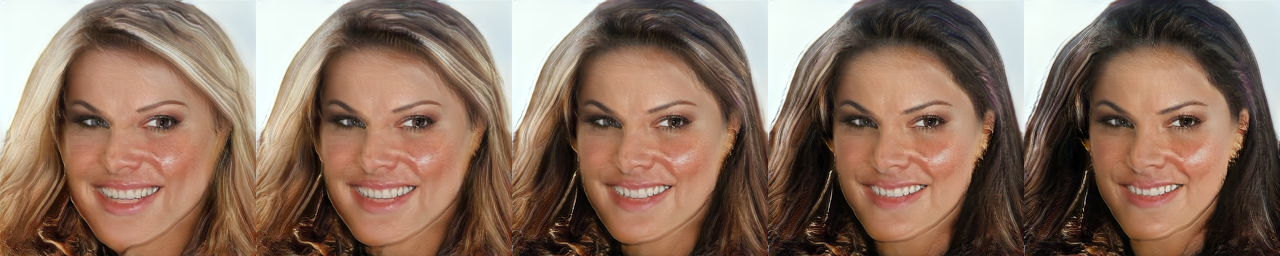}
(a) $+$ Black hair
\end{minipage}
\begin{minipage}{7cm}
\centering
\includegraphics[width=\linewidth]{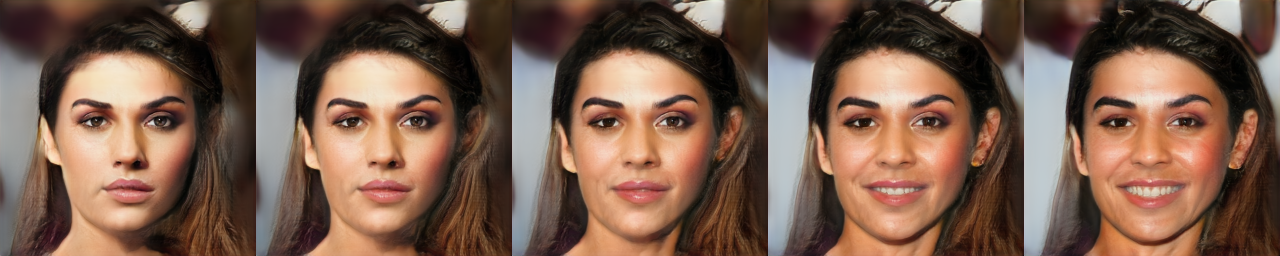}
(b) $+$ Smile
\end{minipage}
}
\vspace{-0.5cm}
\caption{\textbf{Continuous editing results on PGGAN}~\citep{karras2017progressive} with the ``black hair" (left) and the ``smile" (right) attribute.}
\vspace{-0.2cm}
\label{fig:pggan}
\end{figure}

\begin{figure}
	\vspace{-0.2cm}
	\centering
	\includegraphics[width=0.8\linewidth]{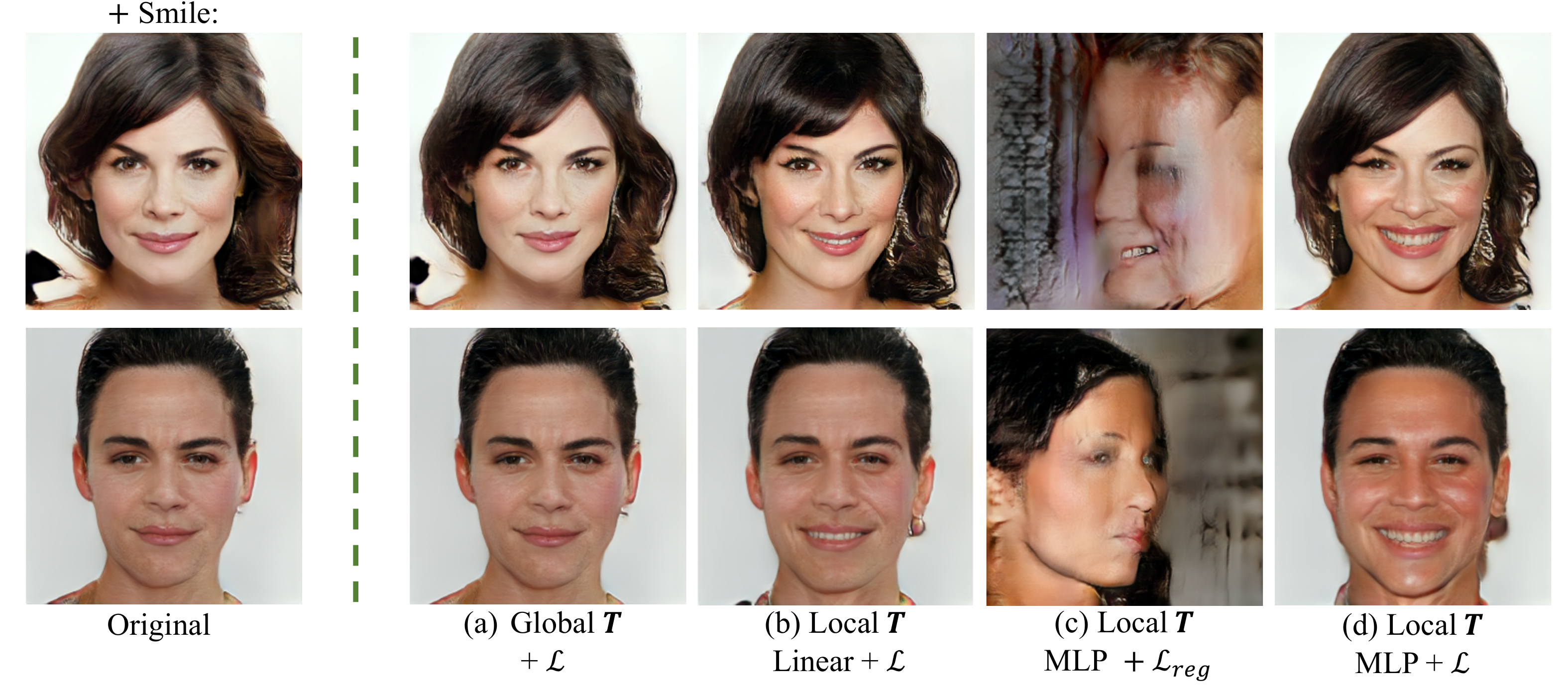}
	\vspace{-0.3cm}
	\caption{\textbf{Comparison of global and local transformations on PGGAN.} Original synthetic images (col.1);  the edited results on enhancing ``smile'' (col.2-5): (a) a global $\bm T$, (b) a local $\bm T$, with linear-layer direction functions, (c) a local $\bm T$ with MLP direction functions trained via $\mathcal{L}_\text{reg}$ loss only, and (d) a local $\bm T$ with MLP direction functions training via the entire loss $\mathcal{L}$. }
	\label{fig:abl}
\end{figure}

\textbf{Manipulation results on PGGAN:} 
For PGGAN, we use local transformations parameterized by three-layer perceptrons with leaky ReLU activations. We show the detailed MLP architecture in Appendix B.1. 
Editing results in Fig.~\ref{fig:pggan} suggest that the proposed approach is able to continuously edit faces with various semantic attributes. 
We also study the differences between local and global directions in Fig.~\ref{fig:abl}. 
Particularly, a global $\bm T$ (in Fig.~\ref{fig:abl} (a)) is a matrix where each column $i$ is given by $\bm d_i$. 
For a local  $\bm T$ (in Fig.~\ref{fig:abl} (b)) $\bm d_i$ is given by a fully-connected layer with identical input and output dimensions. Fig.~\ref{fig:abl} (c) and (d) use the MLP components with the aforementioned structure. We use $\mathcal{L}_{\text{reg}}$ to train $\bm T$ in (c), while the total loss $\mathcal{L}$ is used for (d). The results in Fig.~\ref{fig:abl} indicate that a local $\bm T$, with either linear or MLP structure works well for face attribute edits. Moreover, our loss $\mathcal{L}$ stabilizes the training process of $\bm T$.

\vspace{-0.1cm}
\section{Conclusion}
\vspace{-0.3cm}

We propose an effective approach to semantically edit images by transferring latent vectors towards  meaningful latent space directions. The proposed method enables continuous image manipulations with respect to various attributes. Extensive experiments highlight that the model achieves state-of-the-art performance for targeted image manipulation.

\textbf{Acknowledgements:} This work is supported in part by NSF under Grant No.\ 1718221, 2008387, 1934986 and MRI \#1725729, NIFA award 2020-67021-32799, UIUC, Samsung, Amazon, 3M, and Cisco Systems Inc.\ (Gift Award CG 1377144). We thank Cisco for access to the Arcetri cluster. We thank Amazon for EC2 credits.

\bibliography{main}

\begin{thebibliography}{38}
\providecommand{\natexlab}[1]{#1}
\providecommand{\url}[1]{\texttt{#1}}
\expandafter\ifx\csname urlstyle\endcsname\relax
  \providecommand{\doi}[1]{doi: #1}\else
  \providecommand{\doi}{doi: \begingroup \urlstyle{rm}\Url}\fi

\bibitem[Abdal et~al.(2019)Abdal, Qin, and Wonka]{abdal2019image2stylegan}
Rameen Abdal, Yipeng Qin, and Peter Wonka.
\newblock Image2stylegan: How to embed images into the stylegan latent space?
\newblock In \emph{ICCV}, 2019.

\bibitem[Bhattarai \& Kim(2020)Bhattarai and Kim]{bhattarai2020inducing}
Binod Bhattarai and Tae-Kyun Kim.
\newblock Inducing optimal attribute representations for conditional gans.
\newblock \emph{ECCV}, 2020.

\bibitem[Brock et~al.(2018)Brock, Donahue, and Simonyan]{brock2018large}
Andrew Brock, Jeff Donahue, and Karen Simonyan.
\newblock Large scale gan training for high fidelity natural image synthesis.
\newblock \emph{arXiv}, 2018.

\bibitem[Cao et~al.(2018)Cao, Shen, Xie, et~al.]{Cao18}
Q.~Cao, L.~Shen, W.~Xie, et~al.
\newblock Vggface2: A dataset for recognising faces across pose and age.
\newblock In \emph{IEEE FG}, 2018.

\bibitem[Choi et~al.(2018)Choi, Choi, Kim, et~al.]{choi2018stargan}
Yunjey Choi, Minje Choi, Munyoung Kim, et~al.
\newblock Stargan: Unified generative adversarial networks for multi-domain
  image-to-image translation.
\newblock In \emph{CVPR}, 2018.

\bibitem[Choi et~al.(2020)Choi, Uh, Yoo, et~al.]{choi2019stargan}
Yunjey Choi, Youngjung Uh, Jaejun Yoo, et~al.
\newblock Stargan v2: Diverse image synthesis for multiple domains.
\newblock \emph{CVPR}, 2020.

\bibitem[Goodfellow et~al.(2014)Goodfellow, Pouget-Abadie, Mirza,
  et~al.]{goodfellow2014generative}
Ian Goodfellow, Jean Pouget-Abadie, Mehdi Mirza, et~al.
\newblock Generative adversarial nets.
\newblock In \emph{NeurIPS}, 2014.

\bibitem[H{\"a}rk{\"o}nen et~al.(2020)H{\"a}rk{\"o}nen, Hertzmann, Lehtinen,
  et~al.]{harkonen2020ganspace}
Erik H{\"a}rk{\"o}nen, Aaron Hertzmann, Jaakko Lehtinen, et~al.
\newblock Ganspace: Discovering interpretable gan controls.
\newblock \emph{arXiv}, 2020.

\bibitem[He et~al.(2016)He, Zhang, Ren, and Sun]{he2016deep}
Kaiming He, Xiangyu Zhang, Shaoqing Ren, and Jian Sun.
\newblock Deep residual learning for image recognition.
\newblock In \emph{CVPR}, 2016.

\bibitem[Isola et~al.(2017)Isola, Zhu, Zhou, et~al.]{isola2017image}
Phillip Isola, Jun-Yan Zhu, Tinghui Zhou, et~al.
\newblock Image-to-image translation with conditional adversarial networks.
\newblock In \emph{CVPR}, 2017.

\bibitem[Jahanian et~al.(2019)Jahanian, Chai, and
  Isola]{jahanian2019steerability}
Ali Jahanian, Lucy Chai, and Phillip Isola.
\newblock On the''steerability" of generative adversarial networks.
\newblock \emph{ICLR}, 2019.

\bibitem[Jolicoeur-Martineau(2019)]{jolicoeur2018relativistic}
Alexia Jolicoeur-Martineau.
\newblock The relativistic discriminator: a key element missing from standard
  gan.
\newblock \emph{ICLR}, 2019.

\bibitem[Karras et~al.(2017)Karras, Aila, Laine, et~al.]{karras2017progressive}
Tero Karras, Timo Aila, Samuli Laine, et~al.
\newblock Progressive growing of gans for improved quality, stability, and
  variation.
\newblock \emph{arXiv}, 2017.

\bibitem[Karras et~al.(2019{\natexlab{a}})Karras, Laine, and
  Aila]{karras2019style}
Tero Karras, Samuli Laine, and Timo Aila.
\newblock A style-based generator architecture for generative adversarial
  networks.
\newblock In \emph{CVPR}, 2019{\natexlab{a}}.

\bibitem[Karras et~al.(2019{\natexlab{b}})Karras, Laine, Aittala,
  et~al.]{karras2019analyzing}
Tero Karras, Samuli Laine, Miika Aittala, et~al.
\newblock Analyzing and improving the image quality of stylegan.
\newblock \emph{arXiv}, 2019{\natexlab{b}}.

\bibitem[Kowalski et~al.(2020)Kowalski, Garbin, Estellers, Baltru{\v{s}}aitis,
  Johnson, and Shotton]{kowalski2020config}
Marek Kowalski, Stephan~J Garbin, Virginia Estellers, Tadas Baltru{\v{s}}aitis,
  Matthew Johnson, and Jamie Shotton.
\newblock Config: Controllable neural face image generation.
\newblock \emph{ICML}, 2020.

\bibitem[Laffont et~al.(2014)Laffont, Ren, Tao, et~al.]{laffont2014transient}
Pierre-Yves Laffont, Zhile Ren, Xiaofeng Tao, et~al.
\newblock Transient attributes for high-level understanding and editing of
  outdoor scenes.
\newblock \emph{TOG}, 2014.

\bibitem[Lee et~al.(2020)Lee, Tseng, Mao, et~al.]{lee2020drit++}
H.Y. Lee, H.Y. Tseng, Q.~Mao, et~al.
\newblock Drit++: Diverse image-to-image translation via disentangled
  representations.
\newblock \emph{IJCV}, 2020.

\bibitem[Li et~al.(2019)Li, Qi, Lukasiewicz, and Others]{li2019controllable}
Bowen Li, Xiaojuan Qi, Thomas Lukasiewicz, and Others.
\newblock Controllable text-to-image generation.
\newblock In \emph{NeurIPS}, 2019.

\bibitem[Li et~al.(2017)Li, Fang, Yang, et~al.]{li2017universal}
Y.~Li, C.~Fang, J.~Yang, et~al.
\newblock Universal style transfer via feature transforms.
\newblock In \emph{NeurIPS}, 2017.

\bibitem[Liu et~al.(2018)Liu, Luo, Wang, et~al.]{liu2018large}
Ziwei Liu, Ping Luo, Xiaogang Wang, et~al.
\newblock Large-scale celebfaces attributes (celeba) dataset.
\newblock 2018.

\bibitem[Luan et~al.(2017)Luan, Paris, and Shechtman]{luan2017deep}
F.~Luan, S.~Paris, and E.~andothers Shechtman.
\newblock Deep photo style transfer.
\newblock In \emph{PCVPR}, 2017.

\bibitem[Park et~al.(2019)Park, Liu, Wang, and Zhu]{Park_2019_CVPR}
Taesung Park, Ming-Yu Liu, Ting-Chun Wang, and Jun-Yan Zhu.
\newblock Semantic image synthesis with spatially-adaptive normalization.
\newblock In \emph{CVPR}, 2019.

\bibitem[Plumerault et~al.(2020)Plumerault, Borgne, and
  Hudelot]{plumerault2020controlling}
Antoine Plumerault, Herv{\'e}~Le Borgne, and C{\'e}line Hudelot.
\newblock Controlling generative models with continuous factors of variations.
\newblock \emph{ICLR}, 2020.

\bibitem[Russakovsky et~al.(2015)Russakovsky, Deng, Su,
  et~al.]{russakovsky2015imagenet}
Olga Russakovsky, Jia Deng, Hao Su, et~al.
\newblock Imagenet large scale visual recognition challenge.
\newblock \emph{ICCV}, 2015.

\bibitem[Shen \& Zhou(2021)Shen and Zhou]{shen2020closedform}
Yujun Shen and Bolei Zhou.
\newblock Closed-form factorization of latent semantics in gans.
\newblock \emph{CVPR}, 2021.

\bibitem[Shen et~al.(2019)Shen, Gu, Tang, et~al.]{shen2019interpreting}
Yujun Shen, Jinjin Gu, Xiaoou Tang, et~al.
\newblock Interpreting the latent space of gans for semantic face editing.
\newblock 2019.

\bibitem[Simonyan \& Zisserman(2014)Simonyan and Zisserman]{simonyan2014very}
Karen Simonyan and Andrew Zisserman.
\newblock Very deep convolutional networks for large-scale image recognition.
\newblock \emph{arXiv}, 2014.

\bibitem[Viazovetskyi et~al.(2020)Viazovetskyi, Ivashkin, and
  Kashin]{viazovetskyi2020stylegan2}
Yuri Viazovetskyi, Vladimir Ivashkin, and Evgeny Kashin.
\newblock Stylegan2 distillation for feed-forward image manipulation.
\newblock \emph{arXiv}, 2020.

\bibitem[Voynov \& Babenko(2020)Voynov and Babenko]{voynov2020unsupervised}
Andrey Voynov and Artem Babenko.
\newblock Unsupervised discovery of interpretable directions in the gan latent
  space.
\newblock \emph{ICML}, 2020.

\bibitem[Wang et~al.(2019{\natexlab{a}})Wang, Wang, Zhang, et~al.]{wang2019cnn}
Sheng-Yu Wang, Oliver Wang, Richard Zhang, et~al.
\newblock Cnn-generated images are surprisingly easy to spot... for now.
\newblock \emph{CVPR}, 2019{\natexlab{a}}.

\bibitem[Wang et~al.(2019{\natexlab{b}})Wang, Wang, et~al.]{wang2019detecting}
Sheng-Yu Wang, Oliver Wang, et~al.
\newblock Detecting photoshopped faces by scripting photoshop.
\newblock In \emph{ICCV}, 2019{\natexlab{b}}.

\bibitem[Wang et~al.(2018)Wang, Liu, Zhu, et~al.]{wang2018high}
Ting-Chun Wang, Ming-Yu Liu, Jun-Yan Zhu, et~al.
\newblock High-resolution image synthesis and semantic manipulation with
  conditional gans.
\newblock In \emph{CVPR}, 2018.

\bibitem[Wu et~al.(2019)Wu, Lin, Chang, et~al.]{wu2019relgan}
Po-Wei Wu, Yu-Jing Lin, Che-Han Chang, et~al.
\newblock Relgan: Multi-domain image-to-image translation via relative
  attributes.
\newblock In \emph{ICCV}, 2019.

\bibitem[Zhang et~al.(2018)Zhang, Goodfellow, Metaxas, et~al.]{zhang2018self}
Han Zhang, Ian Goodfellow, Dimitris Metaxas, et~al.
\newblock Self-attention generative adversarial networks.
\newblock \emph{PMLR}, 2018.

\bibitem[Zhou et~al.(2017)Zhou, Lapedriza, Khosla, et~al.]{zhou2017places}
Bolei Zhou, Agata Lapedriza, Aditya Khosla, et~al.
\newblock Places: A 10 million image database for scene recognition.
\newblock \emph{TPAMI}, 2017.

\bibitem[Zhu et~al.(2017{\natexlab{a}})Zhu, Park, Isola,
  et~al.]{zhu2017unpaired}
Jun-Yan Zhu, Taesung Park, Phillip Isola, et~al.
\newblock Unpaired image-to-image translation using cycle-consistent
  adversarial networks.
\newblock In \emph{ICCV}, 2017{\natexlab{a}}.

\bibitem[Zhu et~al.(2017{\natexlab{b}})Zhu, Zhang, Pathak,
  et~al.]{zhu2017toward}
Jun-Yan Zhu, Richard Zhang, Deepak Pathak, et~al.
\newblock Toward multimodal image-to-image translation.
\newblock In \emph{NeurIPS}, 2017{\natexlab{b}}.

\end{thebibliography}
\bibliographystyle{iclr2021_conference}

\appendix
\newpage
\section*{APPENDIX}
\label{app}
\subsection*{A. \hspace{0.5cm}ALGORITHM}
The overall procedure of the proposed method in Algorithm~\ref{alg:1}, taking local transformation $\bm T$ as an example.
\begin{algorithm}
	\caption{Training Procedure}
	\label{alg:1}
	\begin{algorithmic}[1]
		\Input{
		 A pre-trained GAN with $G$, $D$, and input noise distribution $\bm z \sim \mathcal{Z}$; a pre-trained regressor $R$ for predictions on N attributes;
		 an initialized $\bm T$;
		 max iteration number $M$.
		 }
		\For{iteration $m = 1, \dots, M$}
		    \State Sample random noise $\bm z \sim \mathcal{Z}$, and $\bm \eps$
		     \State Compute internal attributes for a synthetic image, i.e., $\bm \alpha = R(G(\bm z))$
		     \State Compute the actual shift value $\bm \delta = CLIP( \bm \alpha +\bm \eps, (0, 1)) - \bm \alpha$
		    \State Compute the transformed latent vector $\bm z' = \bm z + \bm T(\bm z)\bm \delta$ 
		    \State Compute $I' = G(\bm z')$, $\bm \alpha' = \bm \alpha + \bm \delta$
    	        \State Compute attribute predictions $\bm \hat{\bm \alpha}' = R(I')$
    	        \State Compute the loss $\mathcal{L}$
            \State Update $\bm T$
        \EndFor
        \RETURN {$\bm T$} 
	\end{algorithmic}  
\end{algorithm}

\subsection*{B. \hspace{0.5cm}METHOD DETAILS}

\subsubsection*{B.1. \hspace{0.5cm}METHOD ON PGGAN}
\begin{figure}[h!]
\centering
	\includegraphics[width=0.34\linewidth]{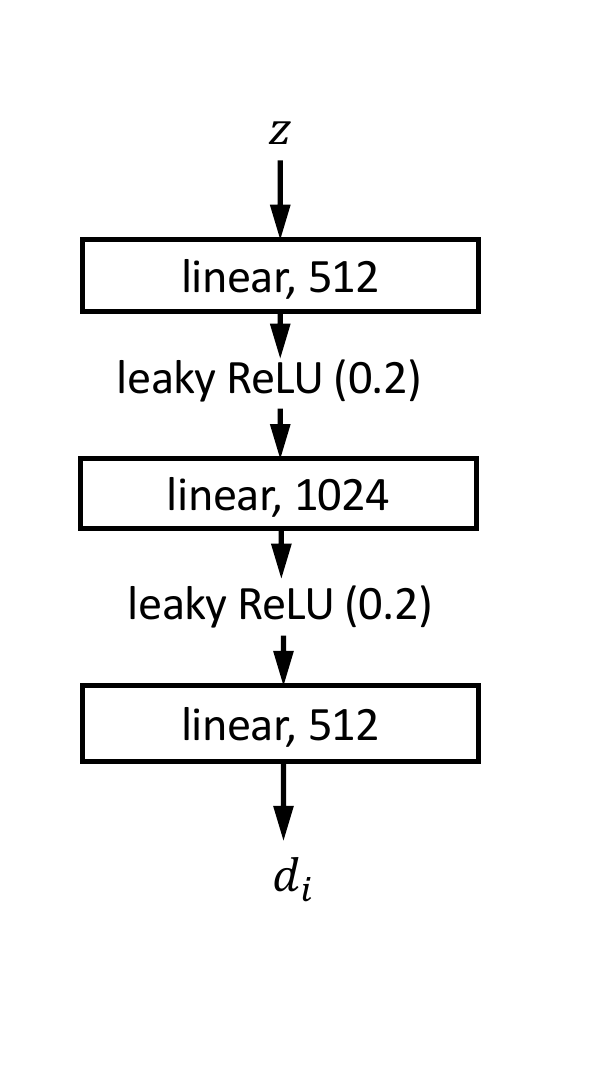}
	\vspace{-1.1cm}
	 \caption{\textbf{Example MLP architecture on PGGAN.}}
	\label{pggan}
\end{figure}

We apply a multi-layer perceptron (MLP) to parameterize a latent space path on PGGAN. Fig.~\ref{pggan} shows the MLP architecture that we use.

In practice, we find that normalizing the direction vector $\bm d_i$ helps preserve image identity during manipulation on PGGAN. Formally, the transformed latent vector can be written as
\begin{equation}
\bm z' = \bm z +  \lambda \bm \eps \frac{\bm d_i}{\Vert \bm d_i\Vert},
\end{equation}
where $\lambda$ is a weight adjusting the direction scale. We set $\lambda=3$.
\subsection*{C.  \hspace{0.5cm} EXPERIMENTAL DETAILS}

\subsection*{C.1 The expectation in $\mathcal{L}_\text{reg}$}
We define the regression loss $\mathcal{L}_\text{reg}$ as 
\begin{equation}
\mathcal{L}_\text{reg} =\quad \mathbb{E}_{\bm z \sim \mathcal{Z}, \bm \eps \sim \mathcal{D}_\eps} 
[-\bm{\hat{\alpha}'} \log{\bm \alpha'} - (\bm 1-\bm{\hat{\alpha}'}) \log{(\bm 1-\bm \alpha')}],
\vspace{-0.1cm}
\end{equation}
 where $\bm \alpha'$ is from the distribution generated by $\bm z \text{ and } \bm \eps$. We now discuss how we derive the distribution of $\bm \alpha'$. To begin, $\bm z$ is sampled from $\mathcal{Z}$ and $\bm \eps$ is sampled from $\mathcal{D}_\eps$. 
 The other variables are generated by $\bm z$ and $\bm \eps$:
 \begin{equation}
 \begin{aligned}
     \bm z’ &= \bm z + \bm T \bm \eps \nonumber\\
     \bm \alpha' &= R(G(\bm z)) + \bm \eps\\
     \hat{\bm \alpha}' &= R(G(\bm z')).
 \end{aligned}
 \end{equation}
 As a result, the distribution of $\bm \alpha'$ is dependent on $\mathcal{Z}$ and $\mathcal{D}_{\eps}$.
\subsection*{C.2 Unsupervised latent-space edits of GANs}
\citet{voynov2020unsupervised} learn a matrix where each column is a direction. Yet, \citet{voynov2020unsupervised} require a human to interpret the learnt directions. 
To avoid bias during the selection of directions in our comparison, we use the pre-trained attribute regressor to automatically identify the most significant directions. Concretely, we first generate 100 images and edit them with  target attributes of various degrees. Next, we use the regressor $R$ on the edited images to predict attributes. We choose for an attribute edit the direction on which all the edited images have the overall highest changing response on the target attribute.

\subsection*{D.  \hspace{0.5cm} Additional Analysis}

\subsection*{D.1 GAN inversion performance}

\begin{figure}[h]
    \centering
    \includegraphics[width=0.8\linewidth]{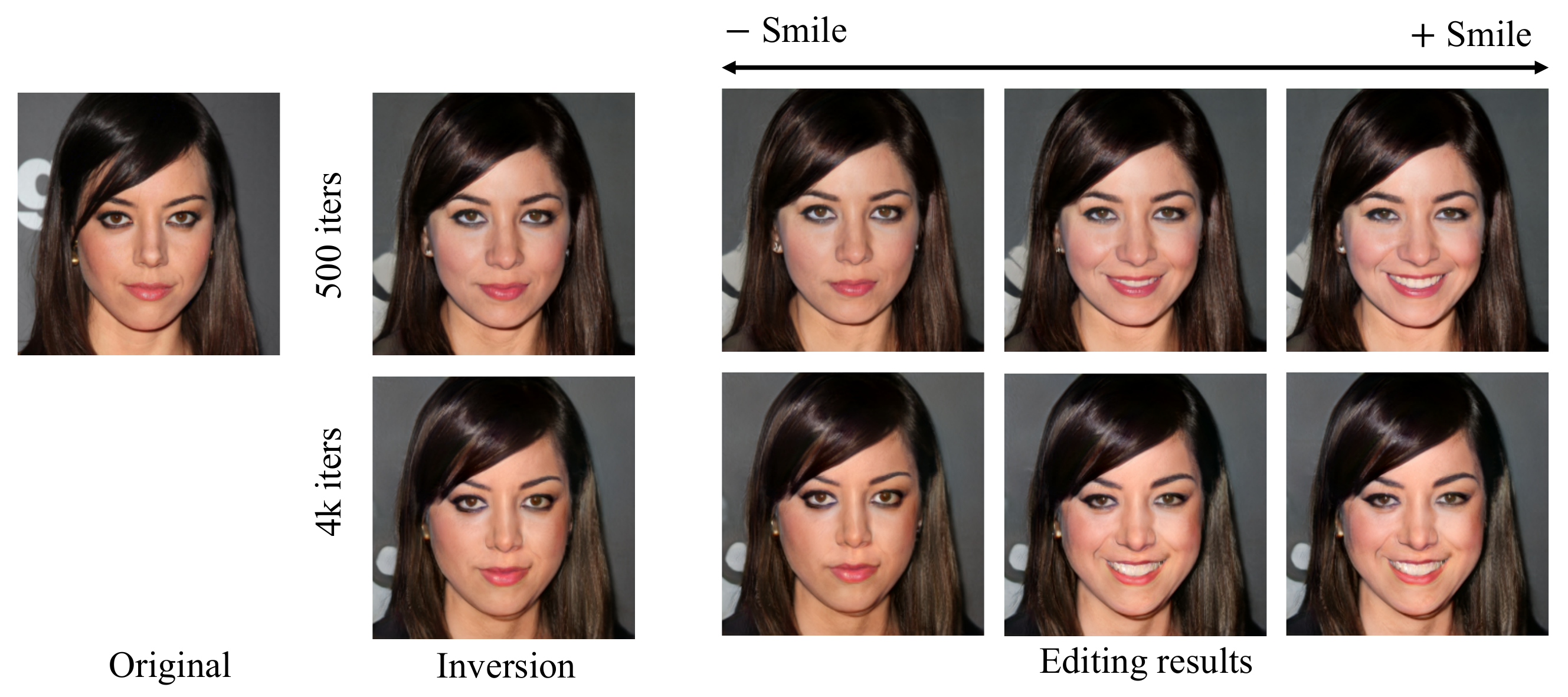}
    \caption{\textbf{Visual comparisons of ``Smile" editing with different reconstructed images}: (col.1) Original image; (col.2) reconstructed images with 500 (top) and 4k (bottom) inversion optimization steps; (col.3-6) results of editing the ``Smile" attribute.}
    \label{fig:sup1}
\end{figure}

We use an optimization-based GAN inversion method~\citep{abdal2019image2stylegan} to find optimal latent vectors that can best reconstruct the real images via the generator.
To examine the effect of GAN inversion performance on our approach, we terminated the GAN inversion approach at different training steps, i.e., 500 and 4,000 iterations. We show averaged reconstruction MSE loss in Tab.~\ref{tab:inversion_mse}. In this case, we reconstructed 20 real face images and edited their ``Smile" and the ``Blond hair" attribute with 10 different degrees $\eps$, i.e., 200 images in total for each attribute editing. Quantitative evaluation on image identity preservation and numerical changes on the other semantically independent attributes for the reconstructed face images are given in Tab.~\ref{tabel:inversion}. The results in Tab.~\ref{tabel:inversion} suggest that the performance of the GAN inversion method affects our method to some degree. Visualized inversion and editing results are shown in Fig.~\ref{fig:sup1}. The qualitative results suggest that our method still works remarkably well on the worse inversion image.

\begin{table}[h]
\caption{\textbf{Averaged MSE loss of reconstructing 20 real face images.} The GAN inversion method~\citep{abdal2019image2stylegan} was trained and terminated at 4k and 500 iterations with averaged MSE loss in the right column.}
\centering
\begin{tabular}{cc}
\toprule
 Training iterations  & MSE\\
 \midrule
 4k iters & 1657.40 \\

 500 iters & 3096.75 \\
\bottomrule
\end{tabular}
\label{tab:inversion_mse}
\end{table}

\begin{table}[h]
\centering
\caption{\textbf{Quantitative evaluation of identity preservation (ID) and attribute changes (Attr)}. We edited two attributes for the real face images, i.e., ``Smile" (col.2-7) and ``Blond hair" (col.8-13). }
\setlength{\tabcolsep}{2.2 pt}{
\begin{tabular}{ccccccc|cccccc}
\toprule
& \multicolumn{6}{c|}{Smile}               
& \multicolumn{6}{c}{Blond hair}   \\
& \multicolumn{3}{c}{ID $(\uparrow)$} 
& \multicolumn{3}{c|}{Attr $(\downarrow)$} 
& \multicolumn{3}{c}{ID $(\uparrow)$} 
& \multicolumn{3}{c}{Attr $(\downarrow)$} 
\\  \hline
$\vert \hat{\eps} \vert$ 
 & {(0, .3]}   & {(.3, .6]}  & {(.6, .9]} 
 & {(0, .3]}   & {(.3, .6]}  & {(.6, .9]}
 & {(0, .3]}   & {(.3, .6]}  & {(.6, .9]} 
 & {(0, .3]}   & {(.3, .6]}  & {(.6, .9]}
\\  \hline
{4k iters} & \textbf{.997} & \textbf{.994} & \textbf{.986}
        & \textbf{.077} & \textbf{.108} & \textbf{.147}
        & \textbf{.988} &\textbf{.979} &\textbf{ .94}
        & \textbf{.1} & \textbf{.112} & \textbf{.16} \\

{500 iters} & .996 & .99 & .982
        & .087 & .138 & .165 
        & .978 & .936 & .914
        & .122 & .158 & .185   \\
\bottomrule
\end{tabular}}
\label{tabel:inversion}
\end{table}

\subsubsection*{D.2 Ablation study}
We conduct an ablation study with single- and joint-distribution training strategies in our method. Specifically, single-distribution sampling refers to learn one attribute direction at a time. In contrast, joint-distribution training means to train multiple attribute directions simultaneously. Fig.~\ref{fig:abl} shows the visualized results with the two training strategies on the face and the scene dataset. We observe that the model with the single-distribution training strategy  generates more unexpected changes as the manipulation degree is getting large, e.g., darker scene colors by the model trained for a single attribute, shown in Fig.~\ref{fig:abl} (b).

\begin{figure}[h]
    \centering
    \begin{minipage}{0.49\linewidth}
    \includegraphics[width=\linewidth]{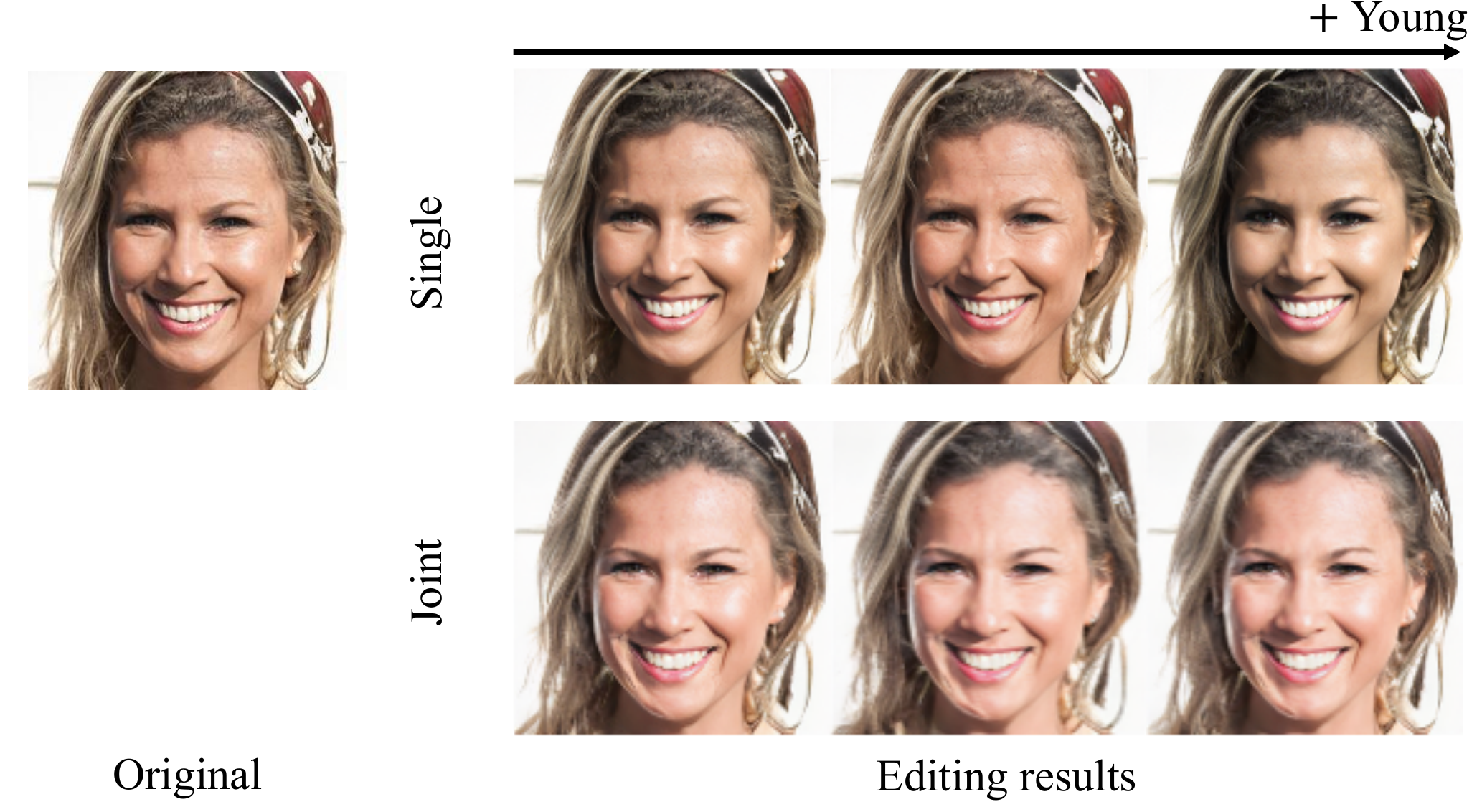}
    \centering
    (a) Visual comparisons of single- and joint-distribution sampling on the face dataset
    \end{minipage}
    \begin{minipage}{0.49\linewidth}
    \includegraphics[width=\linewidth]{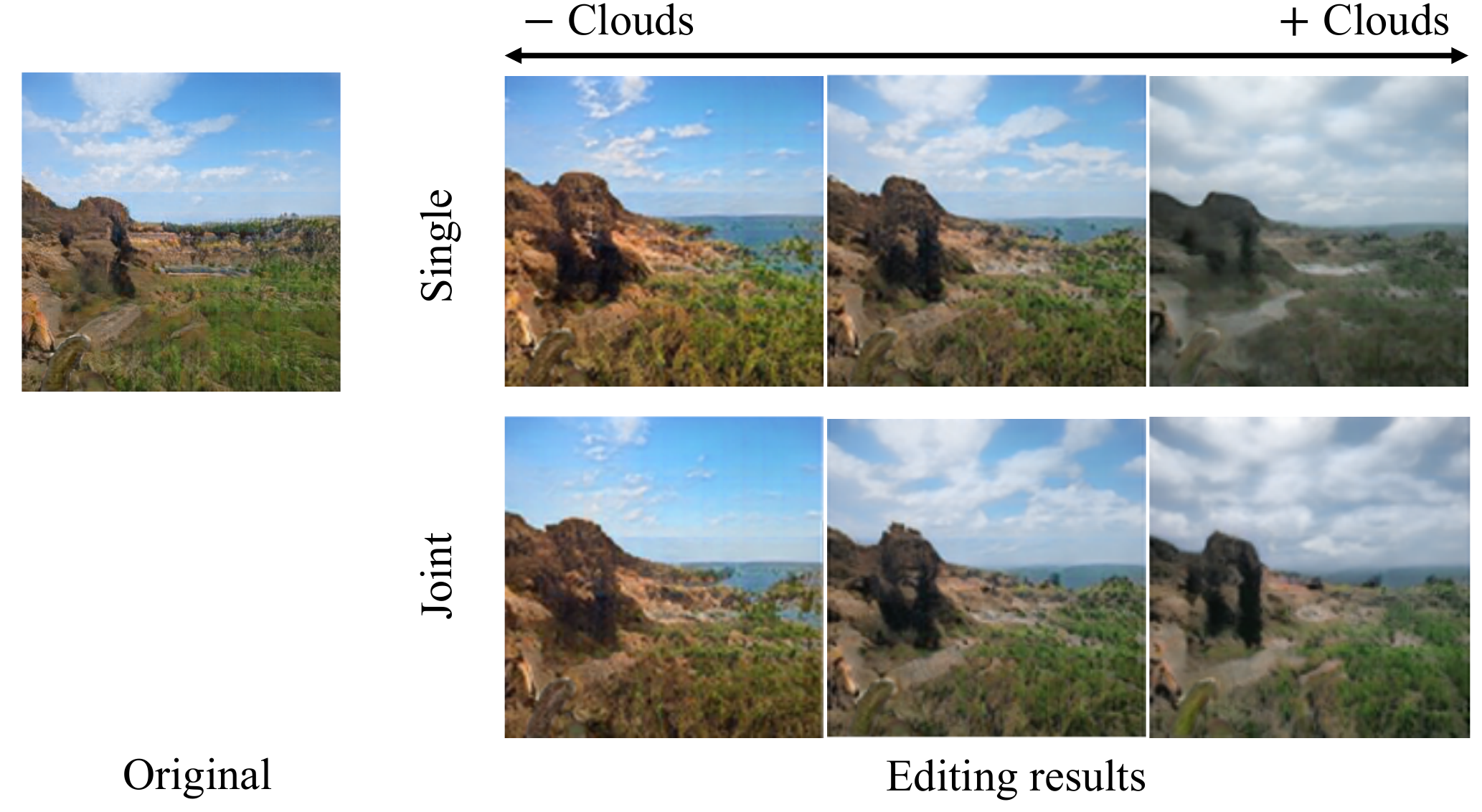}
    \centering
    (b) Visual comparisons of single- and joint-distribution sampling on the scene dataset
    \end{minipage}
    \caption{\textbf{Visual comparisons of single- and joint-distribution training.} Single-distribution training refers to train one attribute direction at a time, while joint-distribution training means to train multiple attribute directions simultaneously.}
    \label{fig:abl}
\end{figure}

\subsection*{E.  \hspace{0.5cm}ADDITIONAL RESULTS}
\subsection*{E.1  \hspace{0.5cm}SCENE IMAGE EDITS ON STYLEGAN2}
\begin{figure}[h]
\centering
    \begin{minipage}{\linewidth} 

\includegraphics[width=\linewidth]{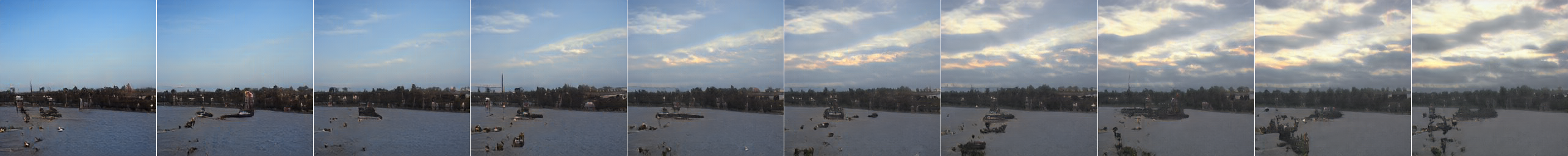}
\includegraphics[width=\linewidth]{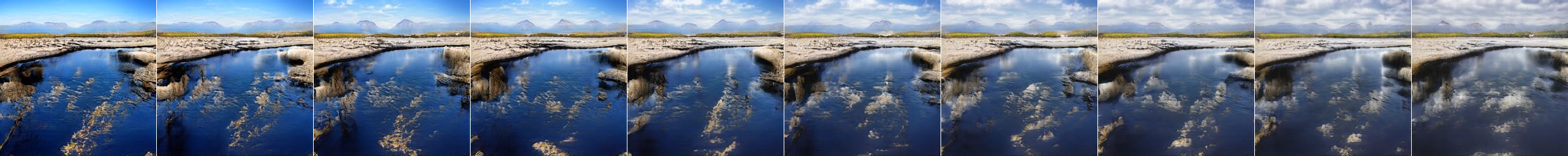}
\includegraphics[width=\linewidth]{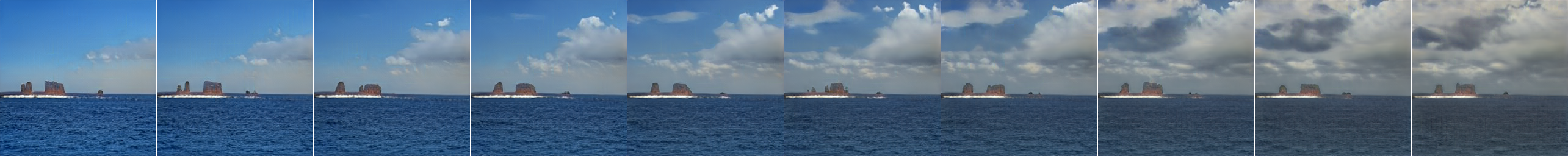}
\centering
\end{minipage}
\caption{\textbf{Additional  results.} Continuous image edits using StyleGAN2~\citep{karras2019analyzing} on the ``clouds'' attribute.}
	\label{fig:clouds}
	\vspace{-0.2cm}
\end{figure}

\begin{figure}
\centering
    \begin{minipage}{\linewidth} 

\includegraphics[width=\linewidth]{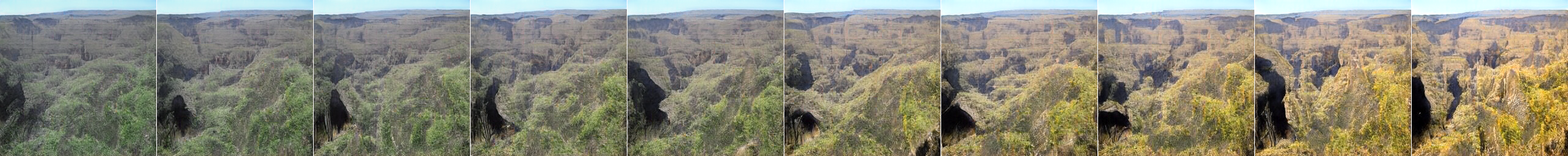}
\includegraphics[width=\linewidth]{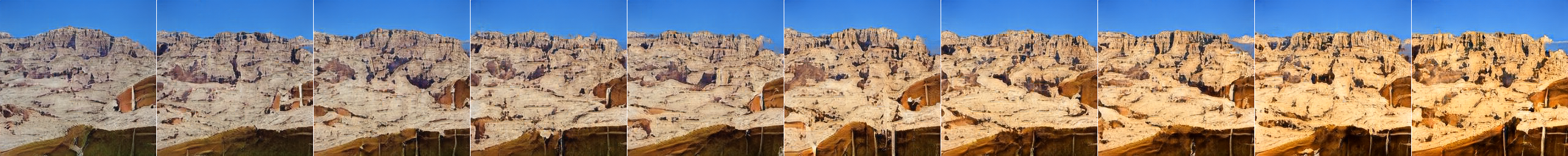}
\includegraphics[width=\linewidth]{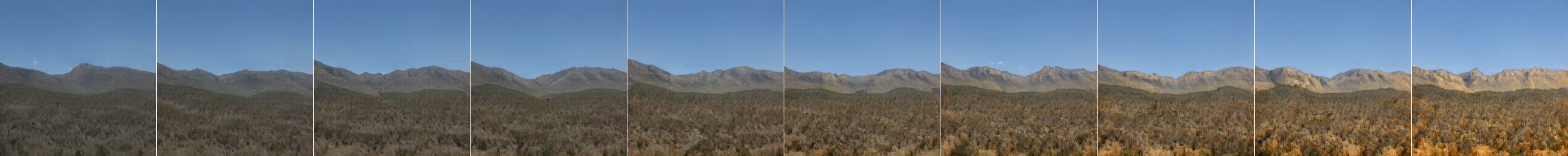}
\centering
\end{minipage}
\caption{\textbf{Additional  results.} Continuous image edits using StyleGAN2~\citep{karras2019analyzing} on the ``brightness'' attribute.}

	\label{fig:bright}
	\vspace{-0.2cm}
\end{figure}

\begin{figure}
\centering
    \begin{minipage}{\linewidth} 
\includegraphics[width=\linewidth]{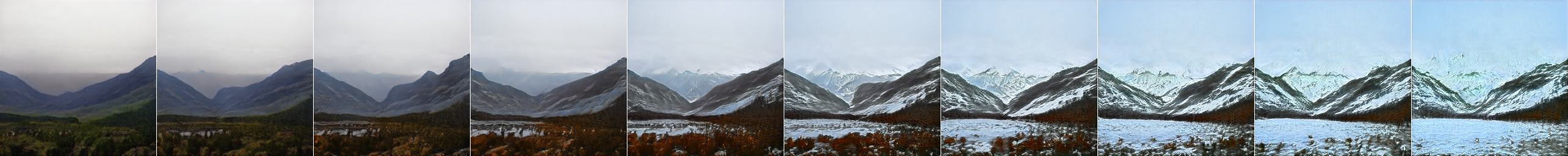}
\includegraphics[width=\linewidth]{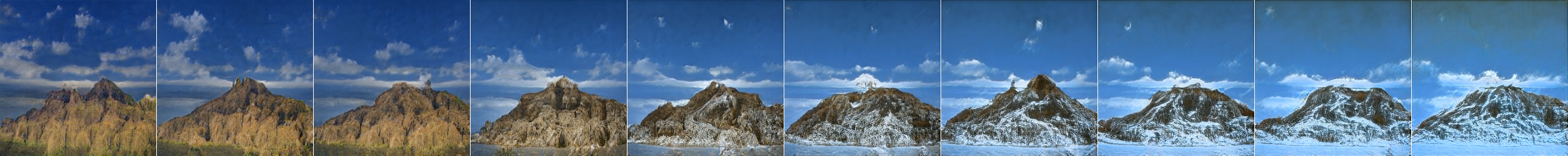}
\includegraphics[width=\linewidth]{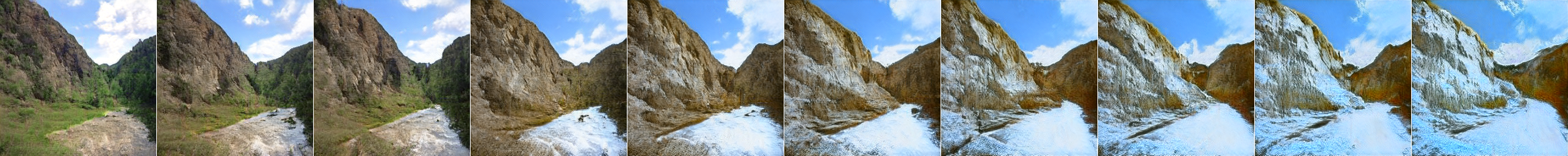}
\centering
\end{minipage}
\caption{\textbf{Additional  results.} Continuous image edits using StyleGAN2~\citep{karras2019analyzing} on the ``snow'' attribute.}

	\label{fig:snow}
	\vspace{-0.2cm}
\end{figure}

\begin{figure}
\centering
    \begin{minipage}{\linewidth} 
\includegraphics[width=\linewidth]{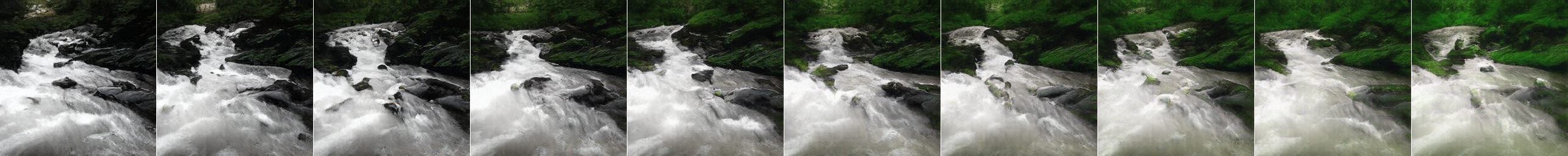}
\includegraphics[width=\linewidth]{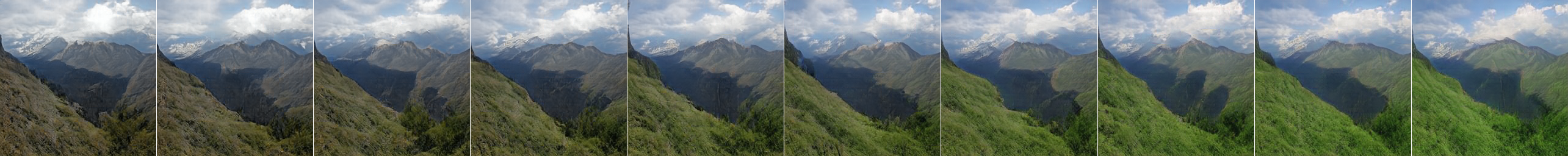}
\includegraphics[width=\linewidth]{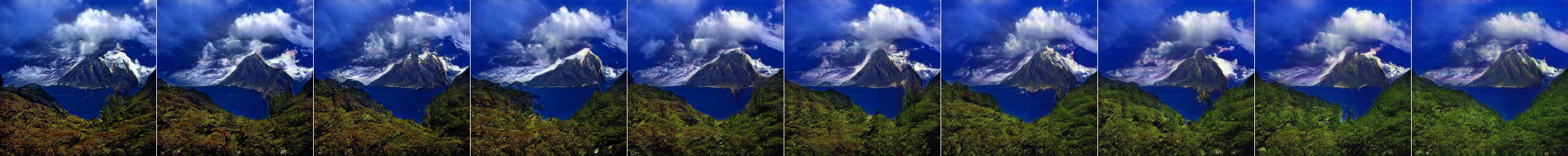}
\centering
\end{minipage}

\caption{\textbf{Additional  results.} Continuous image edits using StyleGAN2~\citep{karras2019analyzing} on the ``summer'' attribute.}

	\label{fig:summer}
\end{figure}

We show more results of our method on continuous scene image edits in Fig.~\ref{fig:clouds} -~\ref{fig:summer}. The semantically edited attributes include ``clouds" (Fig.~\ref{fig:clouds}), ``brightness" (Fig.~\ref{fig:bright}), ``snow" (Fig.~\ref{fig:snow}), and ``summer" (Fig.~\ref{fig:summer}).

\subsection*{F.  \hspace{0.5cm}Broader Impact}
From an application perspective, our method is effective and efficient with regard to image manipulation and photo-realism, which we hope will contribute to 2D and 3D controllable image editing tasks. 
Moreover, usage of deep nets to learn mappings between spaces is still not well understood, 
e.g., mappings between low-dimensional space and image space in classification and generation tasks. We hope our method provides inspiration for representation learning and  a first step for a new view with regard to deep net interpretability. 

Obviously, we are aware of the dangers of automated image manipulation. Similar to deepfake tasks whose aim is to produce fabricated images and videos that appear to be real, improper use of image manipulation approaches  might raise negative issues with regard to information security, property, etc. 
Beyond that, edited image detection techniques~\cite{wang2019detecting, wang2019cnn} are proposed recently to avoid the aforementioned issues,  which promotes growth in both domains.

\end{document}